# From Rule-Based Models to Deep Learning Transformers Architectures for Natural Language Processing and Sign Language Translation Systems: Survey, Taxonomy and Performance Evaluation

Nada Shahin[12] and Leila Ismail[12]

[1] Intelligent Distributed Computing and Systems Lab, Department of Computer Science and Software Engineering, College of Information Technology, United Arab Emirates University, United Arab Emirates
[2] Emirates Center for Mobility Research, United Arab Emirates University, Abu Dhabi, United Arab Emirates

Leila@uaeu.ac.ae

**Abstract.** With the growing Deaf and Hard of Hearing population worldwide and the persistent shortage of certified sign language interpreters, there is a pressing need for an efficient, signs-driven, integrated end-to-end translation system, from sign to gloss to text and vice-versa. There has been a wealth of research on machine translations and related reviews. However, there are few works on sign language machine translation considering the particularity of the language being continuous and dynamic. This paper aims to address this void, providing a retrospective analysis of the temporal evolution of sign language machine translation algorithms and a taxonomy of the Transformers architectures, the most used approach in language translation. We also present the requirements of a real-time Quality-of-Service sign language machine translation system underpinned by accurate deep learning algorithms. We propose future research directions for sign language translation systems.

## Glossary of Terms

**Table 1:** List of abbreviations

| Abbreviation | Definition |
|---|---|
| AR | Augmented Reality |
| ArSL | Arabic Sign Language |
| ASL | American Sign Language |
| CNN | Convolution Neural Networks |
| CSLR | Continuous Sign Language Recognition |
| CTC | Connectionist Temporal Classification |
| DHH | Deaf or Hard of Hearing |
| DNN | Deep Neural Network |
| EBMT | Example-Based MT |
| FFT | Fast Furrier Transform |
| FKNN | Fuzzy-KNN |
| G2S | Gloss-to-Sign |
| G2T | Gloss-to-Text |
| GNN | Graph Neural Networks |
| GPT | Generative Pre-trained Transformers |
| GRU | Gated Recurrent Unit |
| IoT | Internet-of-Things |
| ISLR | Isolated Sign Language Recognition |
| KNN | K-Nearest Neighbors |
| LSTM | Long-Short Term Memory |
| MT | Machine Translation |
| NLP | Natural Language Processing |
| NMT | Neural MT |
| POS | Part-of-Speech |
| QoS | Quality-of-Service |
| RBMT | Rule-Based MT |
| S2G | Sign-to-Gloss |
| S2G | Sign-to-Gloss |
| S2G2T | Sign-to-Gloss-to-Text |
| S2T | Sign-to-Text |
| SLMT | Sign Language Machine Translation |
| SLT | Sign Language Translation |
| SMT | Statistical MT |
| T2G | Text-to-Gloss |
| VR | Virtual Reality |

# 1 Introduction

A Deaf or Hard of Hearing (DHH) person is someone who has hearing loss. According to the World Health Organization, there are approximately 430 million DHH worldwide, which is estimated to double by 2050 [1]. The means of communication for such a person is sign language, a visual language that relies on the movement of different body parts and facial expressions to convey meaning instead of spoken words [2]. Noting that there are 300 sign languages globally [3] that are independent of spoken languages and have their grammar and syntax [4]. To establish communication between the DHH and the broader society, sign language interpreters act as intermediaries to interpret between sign and spoken languages. However, there is a shortage of interpreters. For instance, there are around 10,000 certified interpreters in the United States [5], while there are about 48 million Deaf individuals residing in the country [6]. This scarcity of interpreters, combined with the increasing number of the DHH population, actuates the introduction of real-time automated sign language translation systems. In particular, in the era of smart cities, the population's well-being is essential [7], [8], developing real-time smart translation systems for the DHH is necessary to provide safer, healthier, and more pleasant experiences. The ability to enable communication between DHH and the hearing population could be lifesaving in a disastrous event such as a medical emergency [9]. Consequently, building a software-based real-time translation system that accurately and efficiently translates sign languages to spoken languages and vice versa is crucial.

Machine Translation (MT) was introduced as part of Natural Language Processing (NLP) to translate one spoken language to another [10]. MT algorithms evolved from rule-based to neural networks and are classified into four categories [11]: Rule-Based MT (RBMT), Example-Based MT (EBMT), Statistical MT (SMT), and Neural MT (NMT). Several works exist on machine translation for spoken languages [12], [13]. Despite that MT evolution faced a lot of challenges, including language and context complexity [14], idiomatic expressions [15], and time efficiency [10], there have been a lot of advances in the development of efficient spoken end-to-end translation systems [10]. On the other hand, regarding sign language translation, several studies have applied MT techniques to convert sign language into spoken language and vice versa. These efforts typically concentrate on translating discrete components, either from sign to gloss [16], where signs are the image/video frames and gloss are the linguistic representations of signs, or from gloss to text [17]. The introduction of gloss contributes to higher precision when translating from sign to text [18], or when generating signs [19]. In addition, several existing surveys explored sign language translations in the literature [20], [21], [22], [23], [24]. However, they tackled particular aspects of the sign language translations, such as direct translation from signs to spoken language text [20], [21], [24], signs to gloss [20], [22], [23], [24], gloss to spoken language text [20], [21], [22], [23], spoken language text to gloss [20], [21], [22], or gloss to signs [20], [21], [22], [23]. In our survey, we present a holistic approach where sign language translation enables communication between DHH and hearing individuals in a seamless end-to-end framework from sign to gloss to text and backward, presented in terms of stages. Furthermore, we provide an in-depth analysis of the evolution and current state of the MT systems, particularly focusing on sign language translation revealing insights, challenges, and future research directions. Furthermore, we present a Transformer-based translation case study comparing the performances of its different architectures via empirical evaluations.

In this work, we define a Sign Language Machine Translation (SLMT) algorithm as the composition of two parts that we designate as: 1) Sign-to-Gloss (S2G) recognition, and 2) Sign Language Translation (SLT). S2G falls under computer vision and includes converting static and dynamic gestures and movements in sign language video frames into a corresponding form. This part does not consider the grammar and context in the recognition process. On the other hand, SLT involves translating the recognized content from sign language into spoken language and vice versa while considering both languages' grammar and syntax. This part of the system can be divided into the following subparts: Gloss-to-Text (G2T), and Text-to-Gloss (T2G). The former translates the spoken language text into sign language gloss, a linguistic representation of the signs, while the latter reverses this process.

In this paper, we investigate the SLMT algorithms. We explore all fundamental translation directions by providing a comprehensive analysis of the state-of-the-art, offering insights into the research in this domain. We aim to provide a valuable resource for researchers and developers building deep learning-based end-to-end sign language translation systems. Our main contributions are as follows:

- We present a taxonomy for the sign language components in terms of detection types, sign categories, grammar components, and their interactions.
- We propose a conversational end-to-end SLMT framework that includes the stages involved in the translation process, providing a systematic approach for SLMT researchers and developers to examine translation technologies.
- We compare the SLMT public datasets in terms of size, resolution, and available features such as video, text, and gloss.
- We provide a retrospective analysis of the temporal evolution of SLMT architectures, intending to uncover the challenges that led to the development of the underlying algorithms.
- We classify the SLMT transformer architectures based on the temporal evolution of its functionalities.
- We conduct empirical evaluations of the different transformer architectures in a G2T sign language translation scenario.
- We create a medical-related ASL dataset to perform our empirical evaluations of G2T sign language translation.

The rest of the paper is organized as follows. Section 2 studies the related surveys for sign language translation. Section 3 provides a background on sign language and a taxonomy. Section 4 discusses the machine translation evolution. Section 5 presents the sign language translation framework. Section 6 provides the public SLMT datasets. Section 7 discusses the performance evaluation metrics. SLMT algorithms and taxonomy are presented in Section 8. Section 9 presents the taxonomy of transformer architecture. A case study of translating gloss to text is presented in Section 10. Section 11 presents the challenges and proposed solutions. Section 12 proposes future research directions. Lastly, Section 13 concludes the paper.

## 2 Related Surveys for Sign Language Translation

Several reviews on SLMT exist in the literature [20], [21], [22], [23], [24]. [20], covering the period from 2018 to 2021 explores the advantages, limitations, and challenges of the different methods. It also analyzes these methods based on the performance achieved in the literature. Moreover, it presents sign language datasets, covering both isolated (I) and continuous (C) sign language translation. However, it does not delve into machine translation evolution or provide taxonomies for sign language applied algorithms or transformer architectures. Nevertheless, this survey includes discussions on T2G, gloss-to-sign (G2S), S2G, and G2T translations, but omits sign-to-gloss-to-text (S2G2T) and sign-to-text (S2T) translations. [21] spans a broader period from 2016 to 2022, providing analysis of machine translation evolution, sign language datasets, and continuous sign language data. It also explores the different deep-learning algorithms applied in sign language translation while omitting S2G and S2G2T translations. Nevertheless, it provides partial performance analysis for the sign language translation works, focusing on those post-2018. Despite its comprehensive analysis, this study lacks providing any taxonomies. [22] examines works from 2015 to 2020, exploring sign language datasets (both isolated and continuous) and presenting taxonomies for sign language and the applied algorithms. However, it does not cover machine translation evolution or transformer architectures. It includes discussions and performance analysis for T2G, G2S, S2G, and G2T translations but does not address S2G2T or S2T translations. [23] provides a broader view of the literature from the 1990s to 2020, addressing machine translation evolution, sign language datasets, and performance analysis of the isolated and continuous sign language translation, including T2G, G2S, S2G, and G2T translations. However, it does not cover S2G2T and S2T translations, nor provide any related taxonomies. [24] presents a comprehensive survey from the 1990s to 2021, exploring sign language acquisition, recognition, translation, and linguistic structures, providing a sign language taxonomy. However, it does not delve into the algorithmic approaches applied in the literature nor provides a related taxonomy. Nevertheless, despite providing performance analysis for the literature, its translation coverage is limited, addressing only G2S and S2T translations.

On the other hand, our survey, spanning from 2016 to 2023, represents a unique and comprehensive effort in exploring SLMT. We delve into critical aspects such as MT evolution and sign language datasets for isolated and continuous signs. We investigate both machine-learning and deep-learning algorithms and provide a thorough performance evaluation. What sets our work apart is our discussion of the architectural components in the literature and our presentation of detailed taxonomies for both sign language and the algorithms applied in SLMT. Importantly, we cover all aspects of the translation process, from T2G to G2T, including intermediate and combined processes like S2G, G2S, and the full S2T translation, supported by an in-depth exploration and taxonomy of transformer architectures. In addition, unlike previous surveys, our survey provides empirical evaluations of the different transformer architectures in a G2T sign language translation scenario. This comprehensive approach underscores our unique contribution to the field, offering insights and classifications not previously provided, as detailed in Table 2.

In summary, while existing surveys on SLMT provide valuable insights, they each have limitations in scope and coverage. Our survey addresses these gaps by comprehensively analyzing machine translation evolution, datasets, algorithms, performance metrics, and transformer architectures. We provide detailed taxonomies covering all translation processes and conduct empirical evaluations of the transformer architectures for G2T translation using public and private datasets. making our work a significant contribution to the field of SLMT.

**Table 2:** Comparison between sign language translation-related surveys.

| | | Domains of Comparison | | | | | | | | Covered Translation | | | | | |
|---|---|---|---|---|---|---|---|---|---|---|---|---|---|---|---|
| Work | Period covered | Machine Translation Evolution | Datasets | Isolated (I) / Continuous (C) / Both (B) | Performance | Sign Language Taxonomy | Algorithms Taxonomy | Transformer Architectures Taxonomy | Empirical Evaluation | Text-to-Gloss | Gloss-to-Sign | Sign-to-Gloss | Gloss-to-Text | Sign-to-Gloss-to-Text | Sign-to-Text |
| [20] | 2018-2021 | ✗ | ✓ | B | ✗ | ✗ | ✗ | ✗ | ✗ | ✓ | ✓ | ✓ | ✓ | ✗ | ✗ |
| [21] | 2016-2021 | ✓ | ✓ | C | ✗ | ✗ | ✗ | ✗ | ✗ | ✓ | ✓ | ✗ | ✓ | ✗ | ✓ |
| [22] | 2015-2020 | ✗ | ✓ | B | ✓ | ✓ | ✓ | ✗ | ✗ | ✓ | ✓ | ✓ | ✓ | ✗ | ✗ |
| [23] | 1990s-2020 | ✓ | ✗ | B | ✓ | ✗ | ✗ | ✗ | ✗ | ✗ | ✓ | ✓ | ✓ | ✗ | ✗ |
| [24] | 1990s-2021 | ✗ | ✓ | B | ✓ | ✓ | ✗ | ✗ | ✗ | ✗ | ✗ | ✓ | ✗ | ✗ | ✓ |
| This Work | 2016-2023 | ✓ | ✓ | B | ✓ | ✓ | ✓ | ✓ | ✓ | ✓ | ✓ | ✓ | ✓ | ✓ | ✓ |

## 3 Sign Language Taxonomy

Sign language conveys meaning visually and has distinct linguistic properties that differ from spoken languages [4]. Fig. 1 presents our taxonomy of sign language. We classify the signs based on their sign detection type, sign category, and grammar composition.

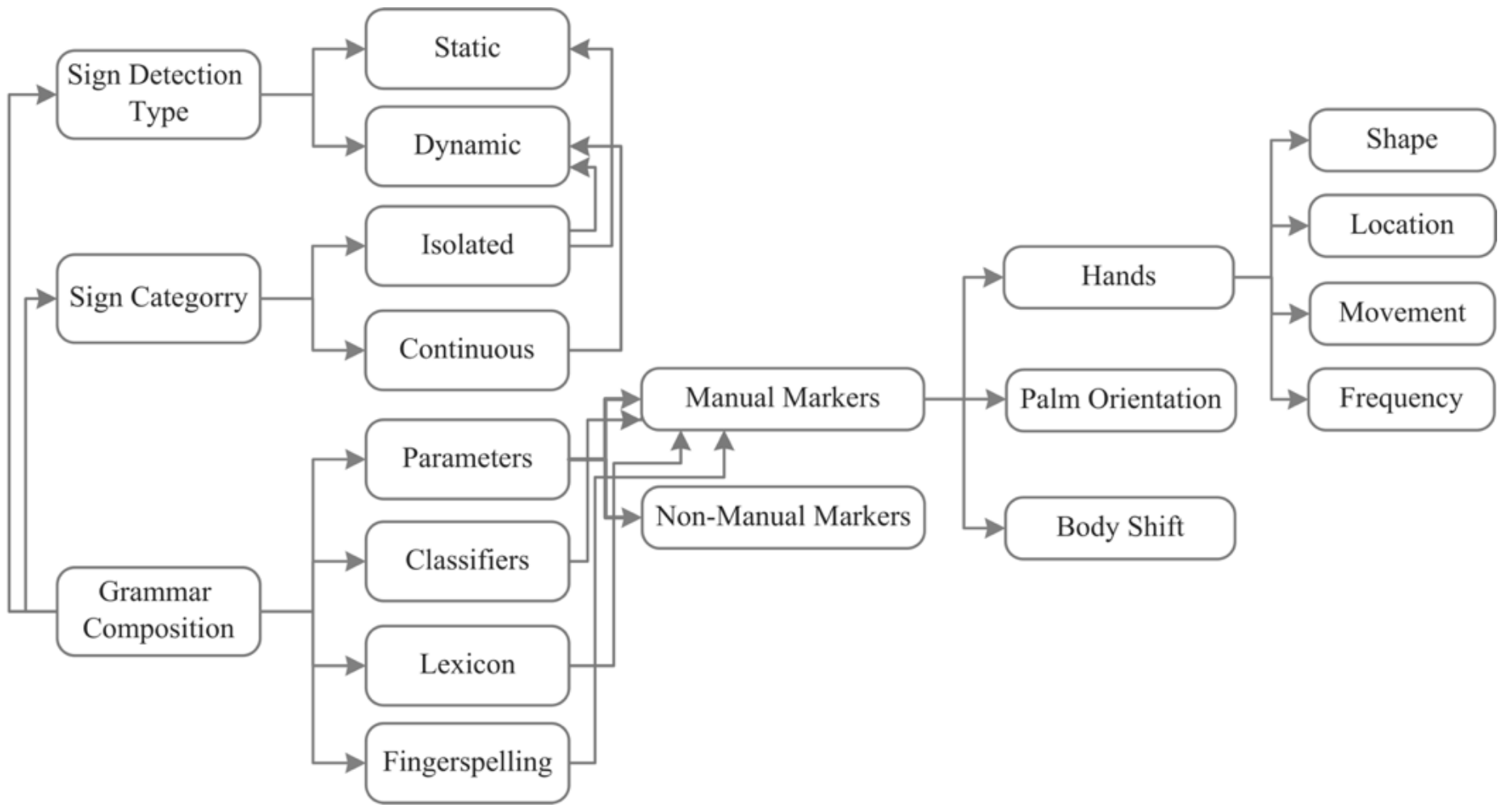

**Fig. 1.** Sign Language taxonomy.

- **Sign Detection.** This category includes the techniques that are used to recognize and interpret sign language. It divides the recognition process into static detection, representing static signs, and dynamic detection, reflecting the multiple frames or videos in which sign language is produced.
- **Sign Category.** We classify signs as isolated recognition, representing words, and continuous recognition, representing sentences. This classification recognizes the varying complexities of sign language communication and how it can adapt to scenarios.
- **Sign Language Grammar.** We categorize the elements that shape the structure and meaning of sign language based on its phonology. Phonology, in any language, is a subfield of linguistics that deals with the study of patterns and phonemes [25]. The phonological parameters of a sign language are categorized into manual and non-manual markers [4]. The manual markers include a) handshape, which refers to the specific shape of the fingers and hands while producing signs. b) handshape movement and frequency, which involves the direction and way the hands and arms move to depict different words and concepts. c) handshape location, where the signs are produced in relation to the body. d) palm orientation refers to the position and direction of the palms when producing a sign. e) body shift, which conveys meaningful messages during conversations and storytelling. Fig. 2 shows an example of these markers. On the other hand, the non-manual markers include body language, eye gaze, and facial expressions such as eyebrows and lip movement. This phonology is crucial in conveying emotions, intensity, and grammatical information. In addition, sign language classifiers are linguistic representations of more than words, allowing the signer to provide more detailed descriptions of objects, people, or events. Fig. 3 shows two different classifiers. Signers utilize classifiers to add depth, context, and specificity to their conversations, making sign language expressive. Moreover, the sign language lexicon refers to the entire set of signs or lexical items to convey meaning. The last grammar element is fingerspelling, which involves manual signs representing the letters of spoken language. It is commonly used to spell words or names for which no established sign exists, such as proper nouns, technical terms, or foreign words [4].

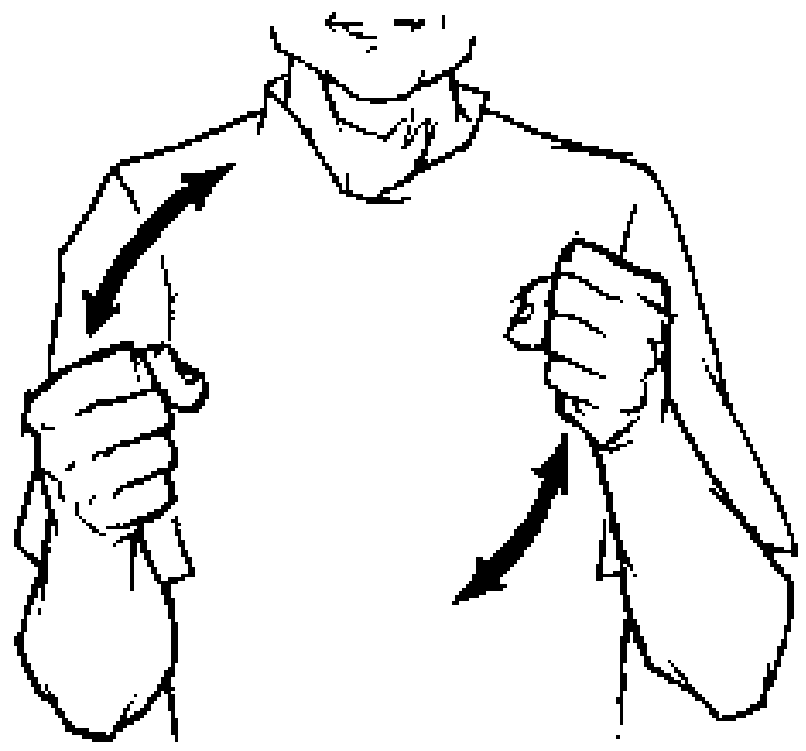

**Fig. 2**. The sign "Drive" in American Sign Language [26]. Closed fists represent the handshape. The movement mimics the motion of turning a steering wheel. The frequency is twice. The handshape location is in front of the body, near the chest. The palm orientation is inwards.

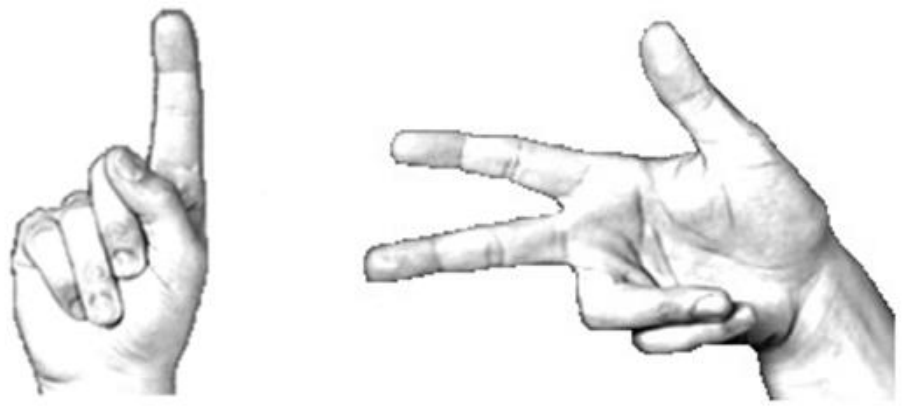

**Fig. 3.** Examples of classifiers. The gesture on the left represents long, thin things, such as people, while the right represents vehicles such as cars [27].

## 4 Machine Translation Evolution

Fig. 4 shows the evolution of MT algorithms over time divided into four categories: Rule-Based Machine Translation (RBMT), Example-Based Machine Translation (EBMT), Statistical Machine Translation (SMT), and Neural Machine Translation (NMT) [28].

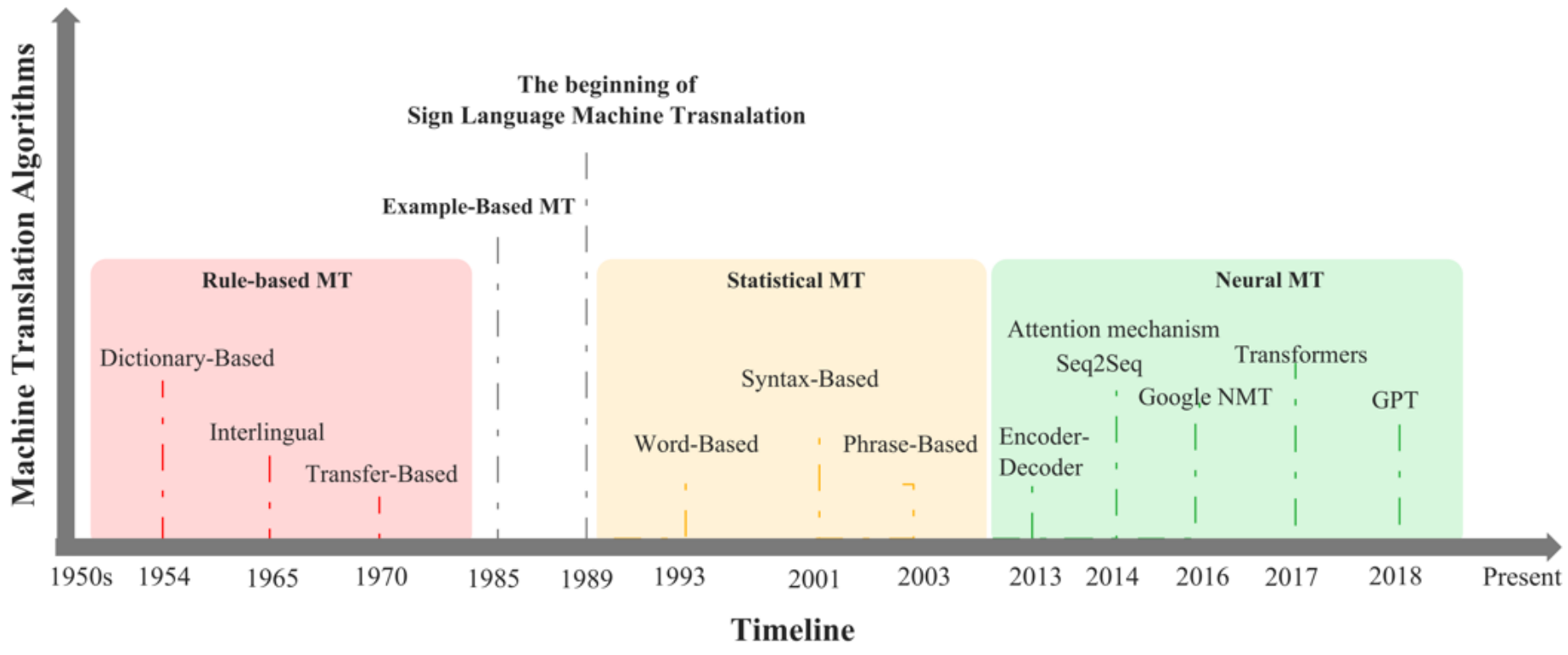

**Fig. 4.** Machine translation algorithms evolution over time.

Machine Translation (MT) began in the mid-20th century due to the world's communication needs. The first documented effort in MT was developing the Automatic Language Processing Advisory Committee project in the United States in 1966 [29]. The early approach relied on linguistic and grammatical rules to translate languages through a rule-based model [30]. However, the model was unsatisfactory due to the languages' complexity. The evolution of MT initially focused on spoken languages. However, it extended to sign languages in the 1980s [21], [31], addressing this visual communication's unique challenges and linguistic nuances. In the following decades, MT research progressed to include statistical and neural approaches in the 1990s and 2000s, respectively, improving translation accuracy [29], [32].

RBMT, introduced in 1954, relies on rules and dictionaries to convert text from one language to another [30]. It consists of three models: dictionary-based, interlingual, and transfer-based [33]. Dictionary-based translation combines dictionaries and grammatical rules to translate text from one language to another by breaking the sentence in the source language into smaller units, such as words or short phrases. The model then searches each unit in a bilingual dictionary to map words or phrases in the source language to their corresponding translations in the target language while applying grammatical rules to maintain proper syntax. Although this approach is straightforward, it falls short in handling the language context [33].

Interlingual MT was introduced to solve the accuracy problem of the dictionary-based MT. This approach identifies the structure and semantics of the text in the source language. Then, it transforms the linguistic meaning into an intermediate language-neutral globally unified representation. The intermediate representation is then used to generate the target language text. This approach raised concerns regarding its efficiency. Therefore, researchers proposed the Transfer-based approach, which functions like the interlingual translation. However, the intermediate representation in the new approach emphasizes structural transfer rather than meaning transfer and is not unified across languages [33].

In contrast to RBMT, EBMT was introduced in the mid-1980s to solve the efficiency and accuracy problems of the RBMT [28]. This was done by utilizing a database of translated sentences for translation reference. The model identifies and modifies examples from the database to generate translations, making this approach effective for handling idiomatic expressions [34].

Furthermore, SMT, introduced in the early 1990s, utilizes probabilistic models and parallel corpora rather than relying on RBMT to produce language translations [35]. The fundamental work by IBM laid the foundational models for SMT, introducing key concepts such as the noisy channel model for learning translation probabilities [36]. Over the years, SMT evolved through several phases, including the introduction of phrase-based models [37], which consider sequences of words or phrases instead of word-for-word translations. Nevertheless, SMT faced challenges in addressing the syntactic differences between languages despite its significant improvements in automatic translation systems, including the ability to rapidly adapt to new languages and domains. These challenges highlighted the need for data-driven approaches that were built based on the groundwork of SMT [28].

More recently, NMT was introduced in 2016 to represent the recent advancement in machine translation [38]. NMT models primarily include encoder-decoder [39], sequence to sequence (Seq2Seq) [40], and attention mechanisms [41]. These models analyze the source text and generate target language output incrementally while calculating translation probabilities to determine the best translation output for each word or phrase. Nevertheless, more recently, NMT expanded to include learning models such as Google NMT [14] and Generative Pre-trained Transformers (GPT) [42]. It is worth noting that GPT has four different versions. GPT-1 [42], released in 2018, was trained on 5 GB of books. Although this model could generate, summarize, complete, and translate text, it suffered from several limitations, such as limited domain-specific knowledge and weak generalization. GPT-2 [43], released in 2019, was trained on 40 GB of Reddit links. This version differs from the previous one by acting as a conversational agent and having a better generalization. Moreover, GPT-3 [44], released in 2020, was trained on a much larger dataset (753 GB) based on Wikipedia, books, academic journals, Reddit links, and common crawl. Lastly, GPT-4 [45], released in 2023, was trained on a 20 TB dataset. Unlike previous versions, which take text only as input, GPT-4 accepts images as input. However, it still outputs text only. In addition, although all versions were trained on data before 2021, GPT-4 can browse the internet through plugins. Recently, [46] conducted experiments using ChatGPT to give insights into its potential for G2T and T2G translations. ChatGPT has promising capabilities in translating from spoken English and Arabic languages to Arabic (ArSL) American (ASL), Australian (AUSLAN), and British (BSL) sign languages and vice-versa.

## 5 Proposed Conversational End-to-End Sign Language Machine Translation Framework

SLMT is a process that involves seamless conversion between sign language and spoken language in both directions, supported by the computer vision and NLP domains. Computer vision involves processing visual data [47], which helps understand and generate signs. NLP [48], on the other hand, contributes to translating these expressions into spoken language. In this section, we present a framework for SLMT in terms of stages to describe the data flow used in the translation process. Fig. 5 illustrates the stages of our framework.

- **Stage 1: Data Collection of Gestures.** Various devices can capture the sign input, including sensor-based wearable devices [49], such as gloves and armbands, and cameras. These cameras can either be vision-based [17] or vision-sensor-based, such as Microsoft Kinect [50], [51]. These instruments can acquire different sign language parameters, such as physical gestures produced by the hands, body movement, and facial expressions, to convey rich linguistic information.
- **Stage 2: Computer Vision for Gesture Recognition.** The signs collected from the previous stage are transformed into gloss, facilitating the S2G process. This builds a connection between gestural communication and linguistic representation, a fundamental element for an accurate translation [52]. Computer vision algorithms can perform this transformation by incorporating data augmentation to enhance the dataset, pre-processing to clean and prepare the data for further analysis, and feature extraction to acquire the glosses for translation [10].
- **Stage 3: Translation to Text.** NLP models translate sign language gloss to spoken language text through G2T translation. This is achieved by applying tokenization and feature extraction methods to comprehend the context and linguistic patterns [53]. Individuals who do not know sign language will be able to understand the output of this translation.
- **Stage 4: Text Acquisition.** Different devices serve as sources for spoken language text input including speech-to-text conversion systems, microphones, or other audio input devices.
- **Stage 5: Translation to Gloss.** This is the reverse direction of the G2T translation, which converts written spoken language text into gloss through T2G via similar steps [54].
- **Stage 6: Sign Generation.** This process focuses on visualizing and generating sign language gestures by translating gloss into sign language expressions represented as image frames facilitating the G2S transformation. The output can be an avatar that functions as a dynamic representation of the signs and facial expressions to close the communication gap between spoken and sign languages [19].

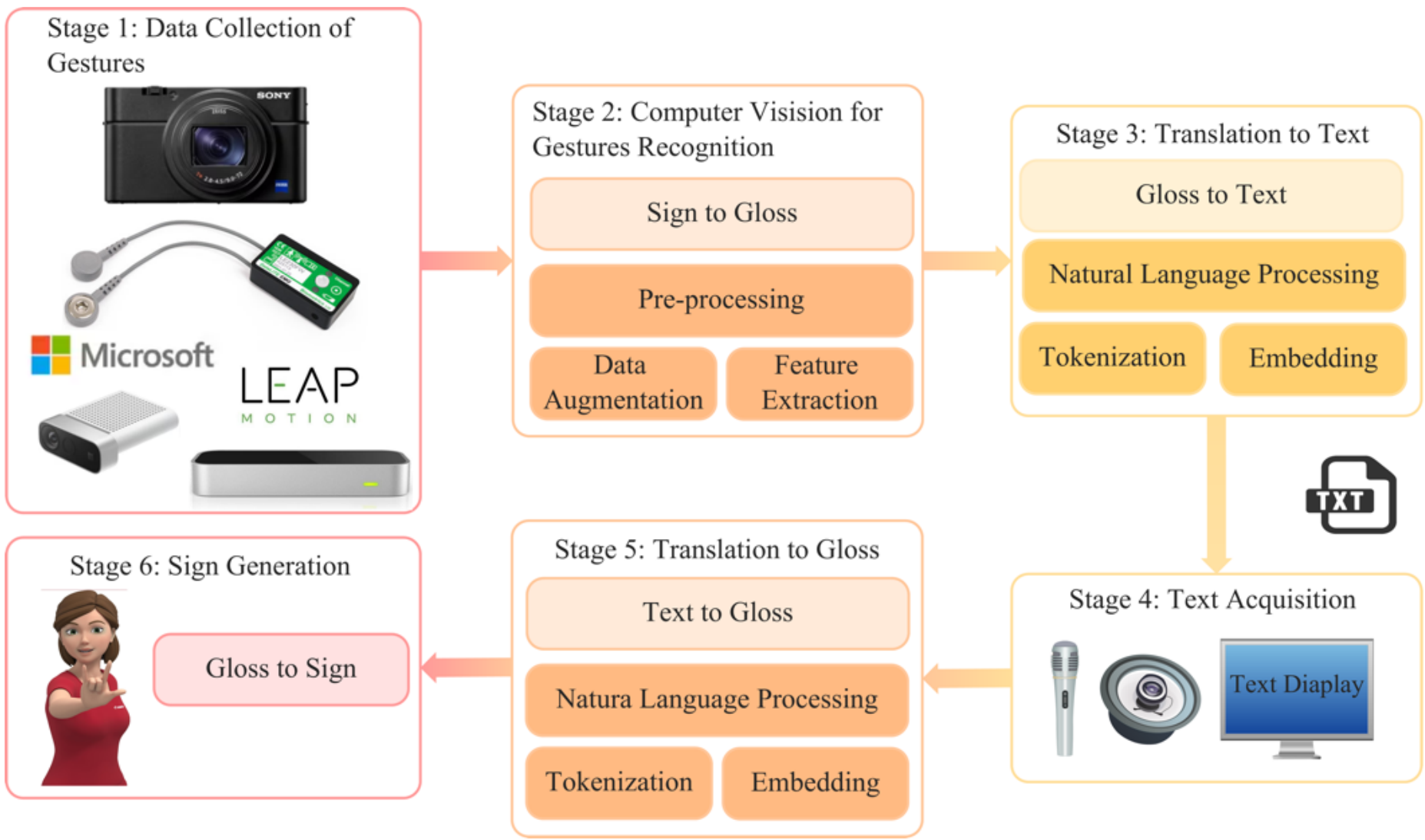

**Fig. 5.** Sign Language Translation Framework.

SLMT systems are mission-critical since they can be utilized during emergencies and disastrous events [9]. Therefore, the following requirements are essential during translation:

- **Large datasets for accurate translation:** SLMT systems require large and diverse datasets to train the AI models and achieve accurate translation effectively. As sign languages exhibit regional variations and individual signing styles [4], a large dataset ensures that the model captures the richness of the signing patterns. In addition, the dataset should include a wide range of vocabulary, expressions, and contexts to enhance the system's ability to translate various sentences with the intended context accurately. An efficient and accurate system must recognize and understand the sign language grammar, facial expressions, and body movements [4]. Achieving high accuracy involves linguistic accuracy and capturing the context embedded in sign languages, which can be distinct from spoken languages.
- **Real-time translation:** Real-time translation is essential for SLMT systems to provide a seamless and uninterrupted user experience [55]. To achieve real-time translation, the SLMT system should satisfy the Quality-of-Service (QoS) requirements such as ultra-low latency and high data rate, for instance in a Metaverse world [56]. This requirement relies on model optimization, efficient computing resources, and lightweight algorithms to ensure rapid and immediate translation [8], [57].
- **Privacy and security measurements**: Language translation often involves personal information being translated. A robust SLMT system must consider the privacy and security standards during a human-MT interaction. This includes the security of user data, encryption during translation, and user consent for any data collection [58], [59]. Compliance with privacy regulations and implementing ethical practices are essential to build trust among users.

## 6 Sign Language Datasets

We present a comprehensive list of the public SLMT in Table 3. Each dataset has essential attributes, including the year, sign language, number of signers (#Signers), number of videos (#Videos), resolution, and acquisition mode. These datasets have different environments, qualities, constraints, and complexities.

SLMT datasets cover different sign languages, including American [60], [61], [62], [63], [64], [65], [66], [67], [68], [69], [70], [71], [72], [73], [74], [75], Arabic [50], [76], [77], [78], [79], Australian [63], Brazilian [80], [81], British [82], [83], Chinese [52], [84], [85], [86], [87], [88], Columbian [89], Finnish [90], French [83], German [17], [83], [91], [92], [93], Greek [83], [94], Indian [95], [96], Irish [97], Korean [98], [99], Persian [100], Polish [101], Russian [102], and Turkish [103], [104]. There is a preference among researchers for developing datasets using Continuous Sign Language Recognition (CSLR) over Isolated Sign Language Recognition (ISLR). This aligns with the practical relevance of CSLR in real-world scenarios, where sign language communication has a continuous and dynamic nature [4]. Moreover, the resolution and acquisition methods vary for some datasets. These variations introduce more diversity and complexity, potentially making them valuable resources for addressing challenging research tasks.

Fig. 6 presents the frequency and average number of videos for the public continuous SLMT datasets. The figure shows that even though American Sign Language (ASL) dominates other sign languages in frequency, it has one of the lowest average numbers of videos (1106). The dominance of ASL in this list is due to its popularity [105]. Hence, the availability of its resources and heightened research interest. On the other hand, the figure shows that Chinese Sign Language has the largest average number of videos (24667). Fig. 7 presents the frequency and average number of videos for the public-isolated SLMT datasets. Like the previous figure, ASL dominates other sign languages in frequency, although it has an average number of videos of 17779, which is the fifth largest sign language. The figure also shows that the Chinese Sign Language has the largest average number of videos (65000). Moreover, Fig. 8 illustrates the coverage of components of the public datasets, showing that most public SLMT datasets cover video frames and gloss, followed by datasets that cover all three components and a few datasets that cover video frames and text. This indicates that the direct S2T translation is not as popular as S2G or S2G2T, highlighting the importance of gloss representation in this process.

In summary, German Sign Language has attracted more attention due to the implementation of Phoenix-2014 [91] dataset and its extension Phoenix-2014T [17]. Based on the popularity of Phoenix-2014T dataset, we suggest utilizing it when introducing new models to ensure accurate comparisons with existing literature until a more extensive and diverse dataset is available. We also suggest creating a large continuous SLMT multilingual dataset to aid in building one generative multilingual SLMT model.

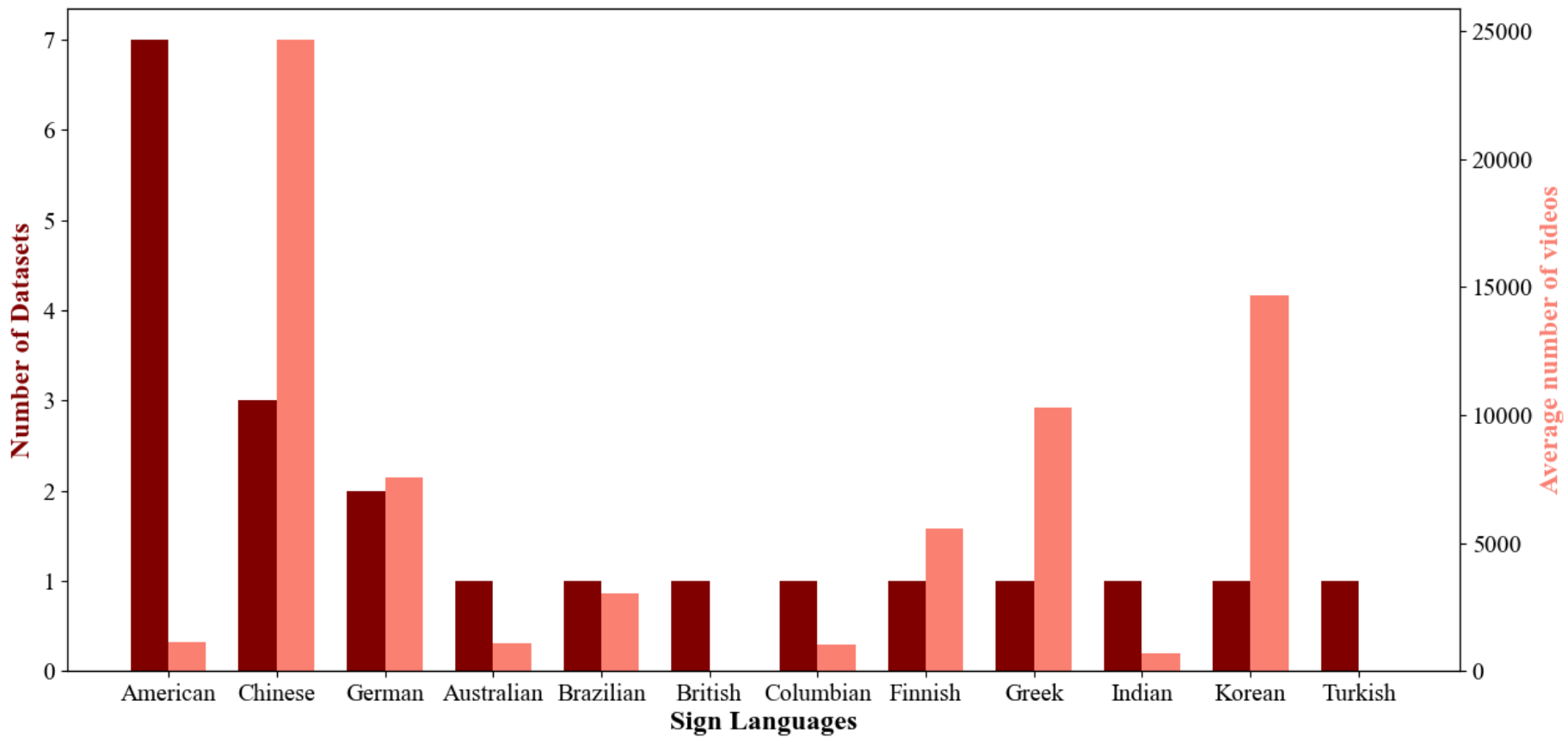

**Fig. 6.** Frequency and average videos for continuous sign languages based on the public datasets.

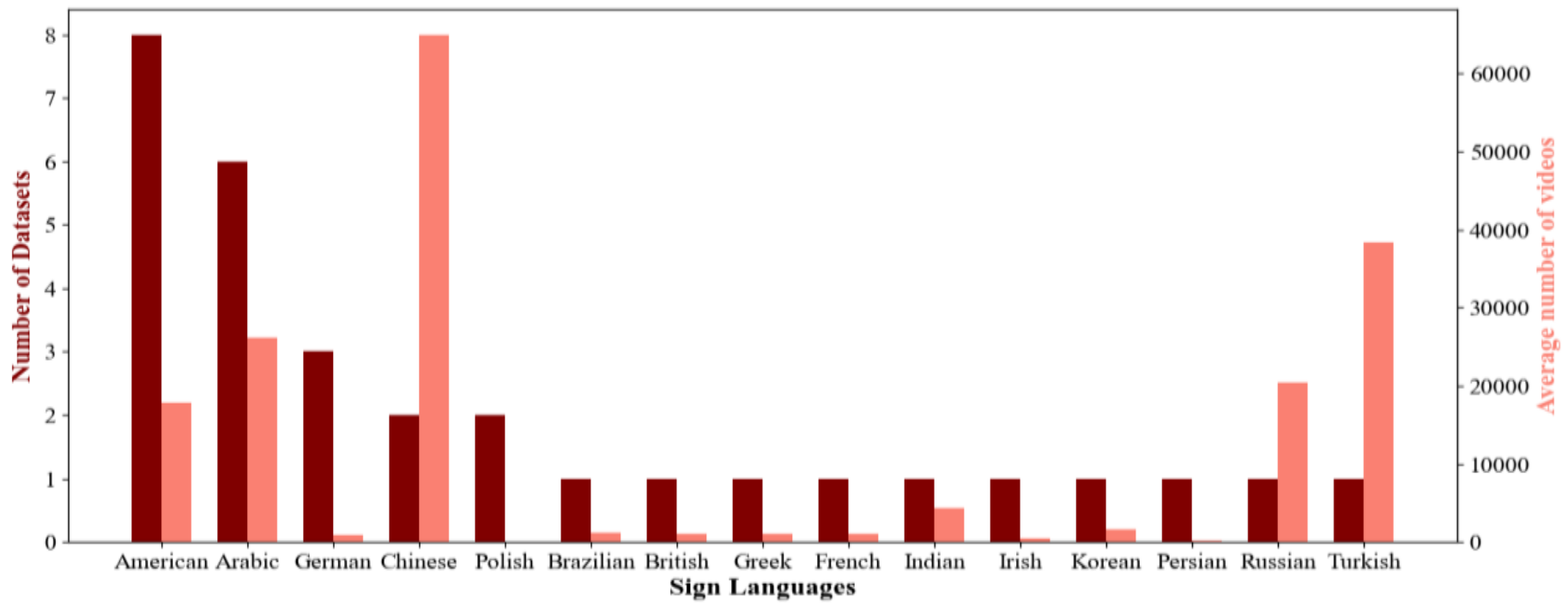

**Fig.7.** Frequency and average videos for isolated sign languages based on the public datasets.

**Table3:** Public sign language datasets.

| Dataset | Year | Sign Language | Video | Gloss | Text | #Signers | #Videos | Resolution | Acquisition | Type |
|---|---|---|---|---|---|---|---|---|---|---|
| BOSTON-50 [60] | 2005 | American | ✓ | ✓ | ✗ | 3 | 483 | 195×165 | RGB | CSLR |
| BOSTON-104 [61] | 2008 | American | ✓ | ✓ | ✗ | 3 | 201 | 321×242 | RGB | |
| ASLLVD [62] | 2008 | American | ✓ | ✓ | ✗ | 4 | NR | Varies | RGB | |
| Auslan dataset [63] | 2010 | Australian | ✓ | ✓ | ✗ | 100 | 1100 | NA | NA | |
| ASLG-PG12 [71] | 2012 | American | ✗ | ✓ | ✓ | NA | NA | NA | NA | |
| BSL Corpus [82] | 2013 | British | ✓ | ✓ | ✓ | 249 | NR | NR | RGB | |
| Devisign [84] | 2014 | Chinese | ✓ | ✓ | ✗ | 8 | 24000 | NR | RGB, Depth | |
| S-pot [90] | 2014 | Finnish | ✓ | ✓ | ✓ | 5 | 5539 | 720×576 | RGB (Betacam) | |
| Phoenix-2014 [91] | 2015 | German | ✓ | ✓ | ✓ | 9 | 6841 | 210×260 | RGB | |
| BosphorusSign [104] | 2016 | Turkish | ✓ | ✓ | ✗ | 10 | NR | 1920x1080 | Kinect v2 | |
| CSL [85] | 2018 | Chinese | ✓ | ✓ | ✗ | 50 | 25000 | 1920×1080 | RGB, depth camera | |
| USTC-CSL [86] | 2018 | Chinese | ✓ | ✓ | ✗ | 50 | 25000 | 1920×1080 | HD RGB | |
| Phoenix-2014T [17] | 2018 | German | ✓ | ✓ | ✓ | 9 | 8257 | 210×260 | RGB | |
| KETI [98] | 2019 | Korean | ✓ | ✓ | ✓ | 10 | 14672 | 1920×1080 | HD RGB | |
| ASLLRP [64] | 2020 | American | ✓ | ✓ | ✓ | 7 | NR | Varies | RGB | |
| CoL-SLTD [89] | 2020 | Columbian | ✓ | ✓ | ✗ | 13 | 1020 | 448×448 | RGB | |
| ASLing [65] | 2021 | American | ✓ | ✗ | ✓ | 7 | 1284 | 450×600 | RGB | |
| How2Sign [66] | 2021 | American | ✓ | ✓ | ✗ | 11 | 2456 | 1280×720 | RGB | |
| LIBRAS-UFOP [80] | 2021 | Brazilian | ✓ | ✓ | ✗ | 5 | 3040 | 640×480 | Kinect | |
| The GSL Dataset [94] | 2021 | Greek | ✓ | ✓ | ✓ | 7 | 10295 | 840×480 | Intel RealSense | |
| ISL-CSLTR [95] | 2021 | Indian | ✓ | ✓ | ✓ | 7 | 700 | NR | RGB | |
| RTWH Fingerspelling [92] | 2006 | German | ✓ | ✓ | ✗ | 20 | 1400 | Varies | Webcam and camcorder | |
| Boston ASL LVD [73] | 2011 | American | ✓ | ✓ | ✗ | 6 | 3300 | 90×90 | NR | |
| ASL Fingerspelling A [74] | 2011 | American | ✓ | ✓ | ✗ | 4 | 48000 | NR | RGB | |
| MSR [75] | 2012 | American | ✓ | ✓ | ✗ | 10 | 12 | 640×480 | Kinect | |
| Dicta-Sign [83] | 2012 | British, German, Greek, and French | ✓ | ✓ | ✗ | 14-16/ SL | +4000 | NA | RGB | |
| DGS Kinect 40 [93] | 2012 | German | ✓ | ✓ | ✗ | 15 | 40 | 640×480 | Kinect | |
| PSL Kinect 30 [101] | 2013 | Polish | ✓ | ✓ | ✗ | 1 | 30 | 640×480 | Kinect | |
| PSL ToF 84 [101] | 2013 | Polish | ✓ | ✓ | ✗ | 1 | 84 | 640×480 | Kinect | |
| Alphabets-Uniform [78] | 2014 | Arabic | ✓ | ✓ | ✗ | 24 | 2800 | NR | NR | |
| Alphabets-Complex [78] | 2014 | Arabic | ✓ | ✓ | ✗ | 8 | 960 | NR | NR | ISLR |
| CSL-500 [88] | 2016 | Chinese | ✓ | ✓ | ✗ | 50 | 125000 | 1920×1080 | RGB, depth camera | |
| SLR-100 [87] | 2016 | Chinese | ✓ | ✓ | ✗ | 50 | 5000 | 1280x720 | Kinect | |
| ISL-HS [97] | 2017 | Irish | ✓ | ✓ | ✗ | 6 | 468 | 640 × 480 | RGB | |
| MS-ASL [67] | 2019 | American | ✓ | ✓ | ✗ | 222 | 25513 | Varies | RGB | |
| ArSL2018 [79] | 2019 | Arabic | ✓ | ✓ | ✗ | 40 | 54049 | 64×64 | Mobile camera | |
| MINDS-Libras [81] | 2019 | Brazilian | ✓ | ✓ | ✗ | 12 | 1200 | 1080×1920 | Kinect | |
| ASL-100-RGBD [68] | 2020 | American | ✓ | ✓ | ✗ | 22 | 100 | 1080×1920 | Kinect | |
| WLASL [69] | 2020 | American | ✓ | ✓ | ✗ | 119 | 21013 | Varies | RGB | |

| | | | | | | | | | |
|---|---|---|---|---|---|---|---|---|---|
| Sign Language Digits [72] | 2020 | American | ✓ | ✓ | ✗ | 218 | 21800 | 100×100 | RGB |
| KSU-SSL [50] | 2020 | Arabic | ✓ | ✓ | ✗ | 40 | 16000 | Varies | RBG and Kinect |
| INCLUDE [96] | 2020 | Indian | ✓ | ✓ | ✗ | 7 | 4287 | 1920×1080 | HD RGB |
| KSL [99] | 2020 | Korean | ✓ | ✓ | ✗ | 20 | 1540 | 1280×720 | RGB |
| RKS-PERSIANSIGN [100] | 2020 | Persian | ✓ | ✓ | ✗ | 10 | 100 | NR | RGB |
| AUTSL [103] | 2020 | Turkish | ✓ | ✓ | ✗ | 43 | 38336 | 512×512 | Kinect |
| KArSL[76] | 2021 | Arabic | ✓ | ✓ | ✗ | 3 | 75300 | Varies | Kinect |
| 27 Class ASL [70] | 2022 | American | ✓ | ✓ | ✗ | 173 | 22490 | 3024×3024 | HD camera |
| AASL [77] | 2023 | Arabic | ✓ | ✓ | ✗ | 200 | 7856 | Varies | RGB |
| Solvo [102] | 2023 | Russian | ✓ | ✓ | ✗ | 194 | 20400 | 1920×1080 | HD RGB |

NA: Not Applicable; NR: Not Reported

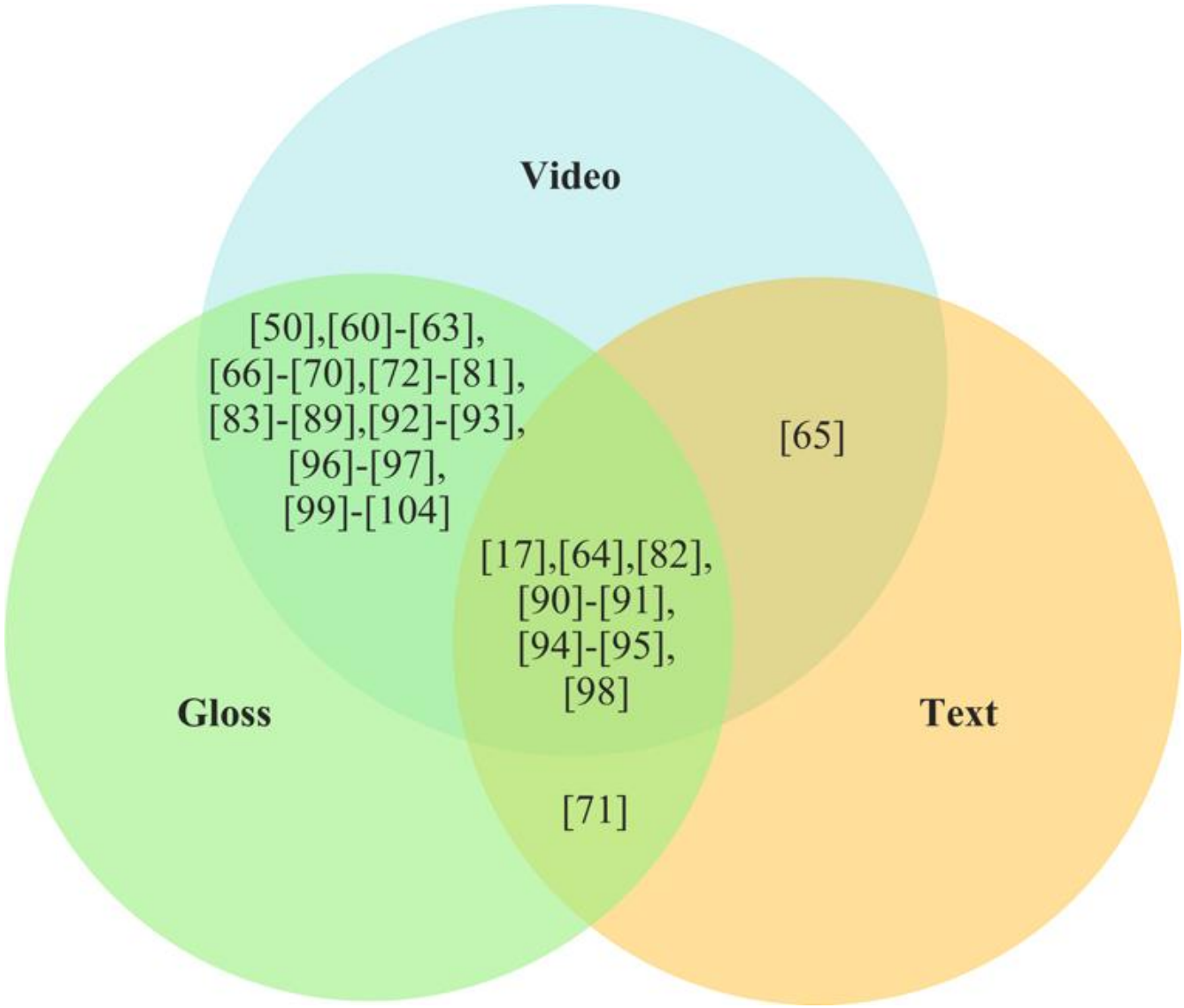

**Fig. 8.** Coverage of components of the public datasets.

# 7 Performance Evaluation Metrics

The assessment of translation systems is essential for understanding their effective-ness and accuracy. This section describes the performance metrics used in SLMT which are Bilingual Evaluation Understudy (BLEU) [106], Recall-Oriented Understudy for Gisting Evaluation (ROUGE) [107], Word Error Rate (WER) [108], and Accuracy.

## 7.1 Bilingual Evaluation Understudy (BLEU)

BLEU is used to measure the similarity between the machine and human translations of sign language to spoken language and vice-versa. It focuses on the precision of n-grams [106]. BLEU scores are expressed on a scale from 0 to 1, where 1 indicates a perfect match with the reference translation.

This metric is divided into BLEU-1, BLEU-2, BLEU-3, and BLEU-4. These variants evaluate the concordance of the respective n-grams between the machine and human translations. This allows the evaluation of the lexical accuracy and appropriateness of the translations.

BLEU score is calculated using Equations (1) and (2).

$$BLEU = BP \cdot e^{\left(\sum_{n=1}^{N} w_n \log(p_n)\right)} \quad (1)$$

where $p_n$ is the precision of n-grams, calculated as the ratio of the number of matching n-grams in the translation to the total number of n-grams in the translation, $w_n$ is the weight of each n-gram size, and $BP$ is the Brevity Penalty.

$$BP = \begin{cases} 1, & if\ c > r \\ e^{(1-\frac{r}{c})}, & if\ c \leq r \end{cases} \tag{2}$$

where $c$ is the length of the candidate machine translation and $r$ is the reference corpus length.

### 7.2 Recall-Oriented Understudy for Gisting Evaluation (ROUGE)

ROUGE assesses the overlap of n-grams, word sequences, and word pairs between the machine and human translations. Unlike BLEU, which primarily focuses on precision, ROUGE gives insight into recall, measuring how many of the human's n-grams are captured by the machine-translated text. This metric helps in understanding the extent to which important information is retained in the translation [107]. ROUGE is calculated using the following Equations (3)-(5).

$$Rouge = 2 \cdot \frac{Rouge_{recall} \cdot Rouge_{precision}}{Rouge_{recall} + Rouge_{precision}} \tag{3}$$

$$Rouge_{recall} = \frac{C}{M} \tag{4}$$

$$Rouge_{precision} = \frac{C}{G} \tag{5}$$

where $C$ is the count of overlapping n-grams between the machine translation and the human translations. $M$ is the total count of n-grams in the human translations. $G$ is the total count of n-grams in the machine translations.

### 7.3 Word Error Rate (WER)

Word Error Rate measures the translation by comparing the machine-translated against the human-translated text on a word-by-word basis [108]. It calculates the number of substitutions (S), deletions (D), and insertions (I) needed to change the machine-translated text into the human-translated text, normalized over the number of words in the human translation (N), as shown in Equation (6).

$$WER = \frac{S+D+I}{N} \tag{6}$$

### 7.4 Accuracy

Accuracy is particularly used for tasks that involve classification, such as identifying specific signs in SLTM systems. It is calculated by Equation (7). However, [49] used Equation (8) to measure the translation accuracy.

$$Accuarcy = \frac{Number\ of\ correct\ translations}{Total\ number\ of\ translations} \tag{7}$$

$$Accuracy = (1 - WER) \times 100 \tag{8}$$

In summary, the evaluation of SLMT systems through a variety of metrics—BLEU, ROUGE, WER, and Accuracy—offers a perspective on system performance, including aspects of precision, recall, error rate, and overall correctness. Together, these metrics enable comprehensive and nuanced assessments of SLMT systems, guiding researchers and developers in identifying strengths, weaknesses, and opportunities for advancements in the field.

## 8 Taxonomy of Sign Language Machine Translation Algorithms

Fig. 9 shows a proposed taxonomy based on a retrospective analysis of the temporal evolution of the SLMT algorithms. We classify these algorithms into four categories: 1) RBMT [54], [109], [110], 2) EBMT [111], 3) SMT [112], [113], [114], [115], and 4) NMT [16], [17], [18], [19], [49], [52], [53], [85], [89], [116], [117], [118], [119], [120], [121], [122], [123], [124], [125], [126], [127], [128], [129], [130], [131], [132], [133], [134], [135]. We map each of these algorithms to the corresponding framework sign language translation stages that were investigated in the literature. We also present a comparison between the SLMT algorithms in the literature in terms of the sign language considered, the dataset(s) used, the translation algorithms and the feature extraction techniques employed, and the corresponding performance, as shown in Table 4.

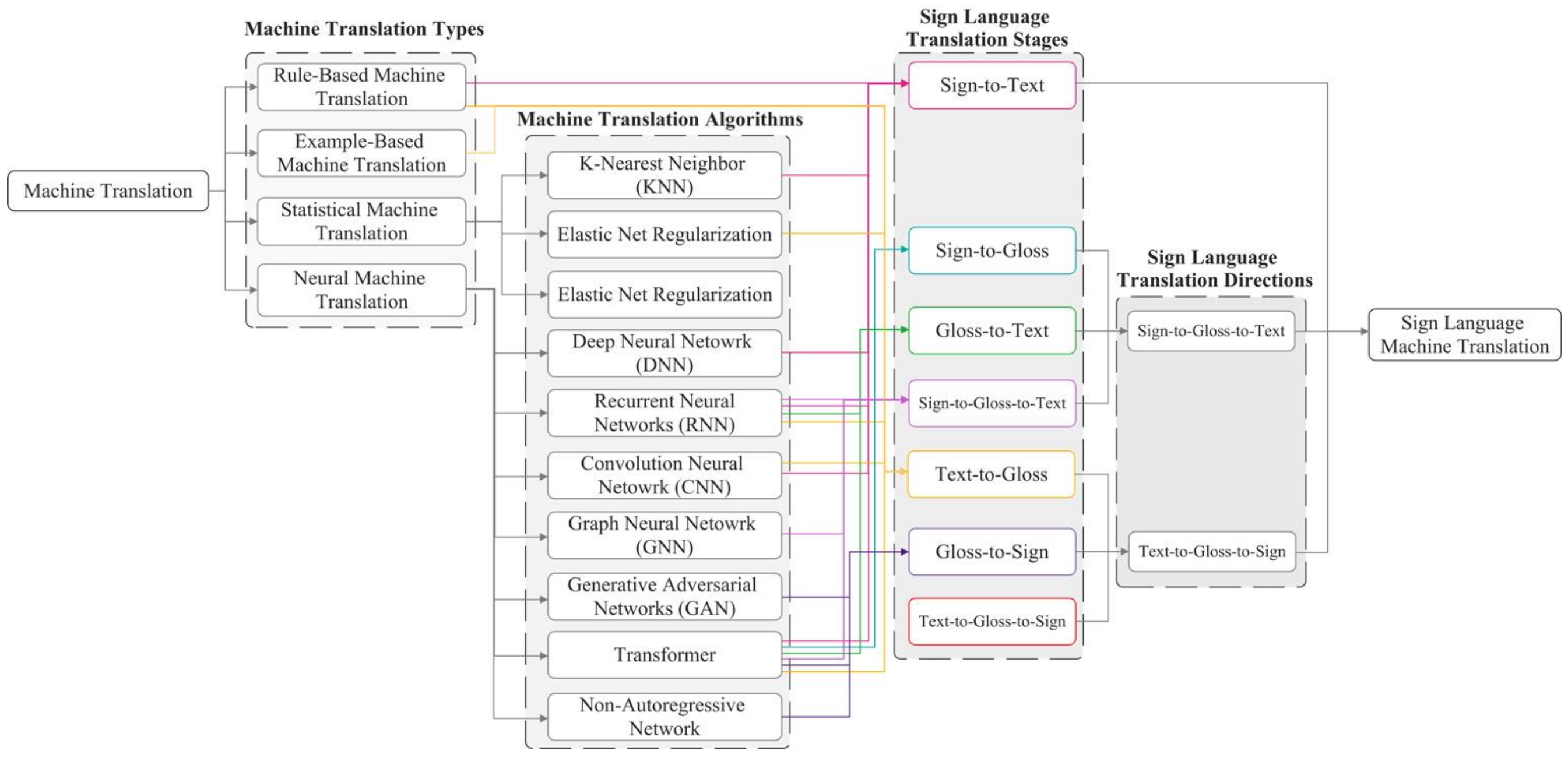

**Fig. 9.** Sign Language Translation Algorithms Taxonomy.

### 8.1 Rule-based Machine Translation (RBMT)

Sign language RBMT relies on grammatical rules, dictionaries, and syntactic and semantic analysis for translation [30]. Few works implemented RBMT for sign language translation. These works are divided into two categories: 1) T2G [54], [109], and 2) S2T [110]. Regarding T2G translation [54], [109], authors followed a three-step rule-based approach which constitutes of Part-of-Speech (POS) tagging, where every word is mapped to its corresponding type, such as noun, verb, and adjectives [136], chunk partial parser, where each sentence is chunked into sub-sentences [137], chunk transfer on the sub-sentence level [54] along with Morpho transfer on the word level [138] to produce the gloss. [54] applied the approach on a Greek dataset while [109] applied the same approach on an Arabic dataset of a smaller size, which resulted in a precision of 84 and 35 respectively. On the other hand, in S2T translation, [110] followed the three-step rule-based approach on Brazilian sign language videos and obtained a precision of 21.1. *Rule1* shows an example of a grammatical rule in sign language [109], where S is the Subject, O is the Object, and V is the Verb.

***IF*** *the input is an Arabic sentence with structure VOS,* ***THEN*** *reorder words to SVO* (Rule1)

In summary, RBMT precision is language-dependent as it is high in Greek and low in Arabic. In addition, the algorithm may face challenges when dealing with dialectical variations of sign languages, expressions, and context-dependency [30].

### 8.2 Example-Based Machine Translation (EBMT)

EBMT was introduced for language translation by looking at the similarity of the corresponding corpus. Despite its precise translations, EBMT has limited translation coverage because no corpus captures all the linguistic nuances [28]. In 2005, [111] presented an example-based approach to translating English text to Dutch gloss by searching for the best matches, based on word occurrences, POS labels, and bilingual dictionaries, and recombining relevant parts of the translated sentence using closed-class words [139] as markers to segment and align source and target sentences. Despite its precision, this method does not scale with a corpus increasing size [140].

### 8.3 Statistical Machine Translation (SMT)

MT models shifted from EBMT to SMT due to the limitations faced by the former type and the need for a more efficient and scalable translation. SMT can solve these issues as it relies on large corpora and probabilistic methods [141] and handles ambiguity better than EBMT. The following are the SMT approaches used in SLMT:

- *Elastic Net Regression*: In 2016, [114] translated 300 ASL sentences using a combination of L1 (Lasso) and L2 (Ridge) penalties of the lasso and ridge method. First, they applied feature mapping to convert the sentences into vectors using the n-spectrum weighted word kernel technique. Then, they mapped between the source and target sentences, which resulted in a linear regression function. This algorithm solves the regression sensitivity and instability and avoids overfitting [142]. Here, L1 penalty penalizes a model based on the sum of the absolute coefficient values. It minimizes the size of all coefficients removing a particular feature of predicted translation from the model. Equation (9) shows the L1 penalty, where $p$ is the number of input features, $j$ is the input feature, and $|\beta_j|$ is the absolute value of the coefficient associated with the $j$th feature.

$$l1 = \sum_{j=0}^{p} |\beta_j| \quad (9)$$

On the other hand, the L2 penalty penalizes models based on the sum of squared coefficient values, as shown in Equation (10). Where $\beta_s^2$ is the square of the coefficient associated with the $j$th feature.

$$l2 = \sum_{j=0}^{p} \beta_j^2 \quad (10)$$

Elastic Net Regularization introduces two hyperparameters: alpha (α) and lambda (λ). Alpha controls the balance between L1 and L2 penalties, with values between 0 and 1, while lambda controls the overall strength of the combined penalty. Equations (11) and (12) present the elastic net penalty and loss formulas, respectively. The loss here is usually a mean square error between the predicted and actual translations.

$$Elastic\ net\ penalty = (\alpha \cdot l1) + ((1 - \alpha) \cdot l2) \quad (11)$$

$$Elastic\ net\ loss = loss + (\lambda \cdot Elastic\ net\ penalty) \quad (12)$$

- *K-Nearest Neighbors (KNN):* In 2017, [113] applied Fuzzy-KNN (FKNN), an extension of KNN, to find the matched sign words in Thai Sign Language and ASL S2T translation. To calculate Fuzzy KNN, the authors used Euclidean Distance [143] in the key point matching process to compare the distance between the two nearest neighbor key points in the dataset and the current key point. Then they calculated the membership value of each data point belonging to multiple words in spoken language using the Gaussian Function [144], as shown in Equation (13). This function describes the degree of belonging of a data point to a particular word in the spoken language. Based on this, the authors assign each sign gesture to a spoken language word, as shown in Equation (14).

$$w_{ij} = exp(-\frac{d_{ij}^2}{2z^2}) \quad (13)$$

where $w_{ij}$is a fuzzification function, and z is a parameter controlling the width of the membership function.

$$FuzzyClass = \frac{\sum_{i=1}^{k} w_{ij} \cdot y_i}{\sum_{i=1}^{k} w_{ij}} \quad (14)$$

where $y_i$ the class label of the $i$-th sample.

- *Phrase-Based Approach*: In 2021, [115] applied the phrase-based approach to translate Marathi text to Indian gloss by applying tokenization and POS tagging, then identifying the phrases' structure using a knowledge analyzer that extracts the syntactical information such as sentence structure and verb tense, which is translated into gloss.

In summary, SMT overcomes the challenges faced by EBMT by relying on large corpora and probabilistic methods, aligning words and phrases across both languages and applying statistical formulas to generate translations [141], thus handling ambiguity better than EBMT. In addition, SMT is more scalable and cost-effective, as it requires fewer manual efforts to maintain the linguistic rules. The SMT methodologies applied in SLMT [112], [113], [114] showcase the different approaches in MT and highlight advancements in linguistic and mathematical techniques.

### 8.4 Neural Machine Translation (NMT)

NMT utilizes deep learning algorithms to translate between languages. Its components include neural network structures, datasets containing original and translated sentences, semantic representations through embeddings, encoder-decoder frameworks, attention mechanisms, and training processes involving backpropagation and gradient descent [40], [41]. These features give superiority to NMT over SMT as they allow the algorithm to handle complex language structures and capture long-range dependencies and language patterns. However, NMT still faces challenges, such as the need for larger datasets and the unique linguistic and spatial characteristics of sign languages.

- *Deep Neural Network (DNN):* In 2017, [123] captured more than 7000 isolated and continuous ASL signs from a private dataset and extracted their features using a leap motion sensor. The authors then translated the signs using a Hierarchical BiDNN with LSTM in S2T translation and obtained a 94.5% translation accuracy. The DNN algorithms they applied consist of neurons, weights, biases, and functions [145]. Initially, the input is fed into the input layer to be represented as vectors. Each neuron performs a weighted sum of inputs in the hidden layers followed by an activation function, as shown in Equation (15). Lastly, the output layer generates the result, depending on the problem.

$$x_i = f(\sum_{j=1}^{N_i}(w_{ij} \cdot a_{i-1j} + b_i)) \quad (15)$$

where $x_i$ is the weighted sum for neuron *i* in layer *i*. $w_{ij}$ is the weight connecting neuron *j* in layer *i-1* to neuron *i* in layer *i*. $a_{i-1j}$ is the output (activation) of neuron *j* in layer *i-1*. $b_i$ is the bias for neuron *i* in layer *i*. *f* is the activation function.

- *Recurrent Neural Networks (RNN):* The implementation of RNN in SLMT focuses on two primary architectures: Gated Recurrent Unit (GRU) and Long-Short-Term Memory (LSTM). Both architectures are well-known for their ability to manage memory through gating mechanisms, allowing for selective retention or forgetting of information. While they share this key principle, GRU and LSTM have major differences in their architectures. GRU involves a hidden state $(h_t)$, reset $(r_t)$ and update ($u_t$) gating mechanisms [146]. $h_t$ indicates the algorithm's memory from the previous time step and is updated at every time step, as shown in Equation (16). $r_t$ controls how much of the previous hidden state ($h_{t-1}$) should be forgotten by using the sigmoid ($\sigma$) activation function to produce values between 0 and 1 for each element in the hidden state, as shown in Equation (17). $u_t$ aims to control how much of the previous hidden state ($h_{t-1}$) should be passed and remembered in the future, as shown in Equation (18). $r_t$ and $u_t$ affect the final translation through the current memory content ($h'_t$), which stores relevant past knowledge, as presented in Equation (19). The final translation output ($o_t$) is calculated by Equation (20).

$$h_t = tanh(w_h * h_{t-1} + w_{hx} * x_t + b_h) \quad (16)$$

where *tanh* is the hyperbolic tangent activation function, $w_h$ is the weight of the current hidden state, $w_{hx}$ is the weight of the input sentence's hidden state, $x_t$ is the vector of the original sentence, and $b_h$ is the bias of the current hidden state.

$$\mathrm{r_t} = \sigma(\mathrm{wr_{h-1}} * \mathrm{h_{t-1}} + \mathrm{wr_x} * \mathrm{x_t}) \quad (17)$$

where σ is the sigmoid activation function, $wr_{h-1}$ and $wr_x$ are the weights of the previous hidden state and the input sentence respectively, in the reset gate.

$$\mathrm{u_t} = \sigma(\mathrm{wu_{h-1}} * \mathrm{h_{t-1}} + \mathrm{wu_x} * \mathrm{x_t}) \quad (18)$$

where $wu_{h-1}$ and $wu_x$ are the weights of the previous hidden state and the input sentence respectively, in the update gate.

$$\mathrm{h'_t} = \tanh(\mathrm{w_x} * \mathrm{x_t} + \mathrm{r_t} \cdot \mathrm{w_{h-1}}) \quad (19)$$

where $\mathrm{w_x}$ is the weight of the input sentence, and $\mathrm{w_{h-1}}$ is the weight of the previous hidden state.

$$\mathrm{o_t} = \mathrm{u_t} \cdot \mathrm{h_{t-1}} + (1 - \mathrm{u_t}) \cdot \mathrm{h'_t} \quad (20)$$

On the other hand, LSTM includes three gating mechanisms: forget ($f_t$), input ($i_t$), and output ($o_t$) gates. $f_t$ removes that information that is no longer useful by taking the vector of the original sentence ($x_t$) and the previous hidden state ($h_{t-1}$) as inputs, as shown in Equation (21). $i_t$ adds valuable information to the cell while $o_t$ controls which information should be used to produce the translation. Equations (22) and (23) represent the input and output gates respectively.

$$f_t = \sigma(wf_{h-1} * h_{t-1} + wf_x * x_t + b_f) \quad (21)$$

$$\mathrm{i_t} = \sigma(\mathrm{wi_{h-1}} * \mathrm{h_{t-1}} + \mathrm{wi_x} * \mathrm{x_t} + \mathrm{b_i}) \quad (22)$$

$$\mathrm{o_t} = \sigma(\mathrm{wo_{h-1}} * \mathrm{h_{t-1}} + \mathrm{wo_x} * \mathrm{x_t} + \mathrm{b_o}) \quad (23)$$

where w is the weight and b is the bias.

The application of GRU and LSTM in SLMT has led to significant developments and findings. For instance, in 2018, [17] applied GRU with a Luong attention mechanism for G2T and S2G2T translation showcasing enhanced performance on PHOENIX-2014T dataset compared to LSTM with an equivalent attention mechanism. GRU was also applied in T2G [117], [118] and S2T [124], [127], [132] translation. In T2G translation, [117] applied the Bahdanau attention and tested it on translating English sentences to Pakistani Sign Language gloss while [118] applied the Luong attention and tested it on PHOENIX-2014T dataset. Nevertheless, in S2T translation, [124], [127] tested their proposed models on PHOENIX-2014T dataset while [132] tested the architecture with the KETI and PHOENIX-2014T datasets. Despite KETI's smaller size, it yielded superior performance, suggesting differences in dataset characteristics or model adaptability need further investigation.

A comparison within the specific contexts of G2T and T2G translations on PHOENIX-2014T dataset illuminated the efficacy of these models, with [118] showing superior results over [17]. We believe that variance in performance stems from the distinct architectural parameters of the models.

Conversely, LSTM's application in SLMT highlights its ability to reduce the vanishing gradient problem and adeptly capture long-term dependencies within sequences, unlike GRUs [147]. This is achieved through its distinct memory cell, setting it apart from GRU's approach. Few works applied LSTM in SLMT. In 2018, [85] proposed a framework consisting of 3D CNN for feature extraction and a hierarchical LSTM encoder-decoder model for S2T translation. The authors tested the model on their proposed CSL dataset. Other works also applied LSTM with varied feature extraction methods in the same translation manner [49], [85], [89], [128]. LSTM was also applied in T2G [116] and S2G2T translations [122].

In conclusion, both GRU and LSTM architectures play essential roles in SLMT. The choice between GRU and LSTM often relies on the specific requirements of the translation task, including the need for memory efficiency, complexity of the sequence dependencies, and computational resources.

- *Convolution Neural Networks (CNN):* CNN is a deep learning architecture used in image processing [148]. It was implemented in sign video frame processing as in S2T translation in 2021 [125]. Here, the researchers applied 3D CNN for feature extraction and translation, utilizing three datasets for three sign languages: Brazilian, Indian, and Korean. [129] proposed a model that consists of CNN with an attention layer and 2D ResNet and tested the model on a private Malaysian Sign Language dataset. The algorithm consists of convolutional, pooling, and fully connected layers. The convolutional layer involves a kernel that slides over the input image and computes a dot product at each location, as shown in Equation (24). On the other hand, the pooling layer reduces the spatial dimensions of the image, as shown in Equation (25). The output from convolution and pooling layers is then passed through fully connected layers to produce the result. These fully connected layers are similar to the ones in DNN.

$$(I * K)(x, y) = \sum_i \sum_j I(x + i, y + i) \cdot K(i, j) \quad (24)$$

$$\mathrm{P}(\mathrm{x}, \mathrm{y}) = \max\big(\mathrm{I}(\mathrm{x}', \mathrm{y}')\big) \text{ for } \mathrm{x}' \in [\mathrm{x}, \mathrm{x} + 2], \mathrm{y}' \in [\mathrm{y}, \mathrm{y} + 2] \quad (25)$$

where$(I * K)(x, y)$ is the result of the convolutional operation at the $x, y$ location. I is the input image, and K is the kernel. $P(x, y)$ is the output of the pooling layer at the location $x, y$.

- *Graph Neural Networks (GNN):* In 2022, [120] implemented an encoder-decoder architecture to convert sign language videos into hierarchical spatiotemporal graph representations. The encoder uses graph convolution and a graph self-attention method to represent the sign videos in latent space, followed by hierarchical graph pooling. The decoder generates text using encoded reorientations in a hierarchical GNN. The algorithm captures the relationships between words or phrases. Here, the source and target sentences are represented in two different graphs. The words or phrases are the nodes, and the relationships are represented in edges [149].
- *Generative Adversarial Networks (GAN):* In 2020, [118] applied GAN to produce sign videos from gloss input and evaluated the accuracy of the generated videos using qualitative measures. The generated videos' resolution and expressiveness were affected by the low resolution of the training data. [135] proposed a Dynamic GAN model to produce high-quality realistic sign language videos. The model consists of three main stages: 1) A GAN network to generate human-based sign gesture videos based on the provided gloss and ground truth images. 2) A CNN network to improve the quality of the generated samples by applying image alignment techniques. 3) A discriminator network to evaluate the quality of the generated. The model produced visually coherent and contextually relevant sign language gestures with high accuracy.
- *Transformers:* In 2020, [16] applied the transformer encoder-decoder architecture in S2G2T translation. Other works later applied the same architecture for the same translation [18], [52], [53], [119], [121]. [16], [52], [121], and tested the algorithm on PHOENIX-2014T dataset while [18], [53], [119] tested it on CSL-Daily dataset. [52] and [119] outperformed the other works in each category, while [119] surpassed [52]. We believe that this is due to the architecture parameters and the dataset size. In terms of S2G translation, [16] implemented an encoder-only transformer trained using a Connectionist Temporal Classification (CTC) loss to predict the gloss sequences. [119] adopted a Self-Mutual Knowledge Distillation model based on CNN to extract the visual features of the sign videos and convert them to 1D temporal convolution, which in turn they use as an input to the encoder-decoder transformer to translate the sign videos into glosses. Both works tested the transformer model using PHOENIX-2014T dataset. When compared, [119] outperforms [16] as it has a lower word error rate. On the other hand, in the G2T translation, [19] implemented an encoder-decoder transformer model variation that we call "encoder-decoder transformer fusion." This variation includes an FFT block and a gated bilateral fusion. However, the researchers omitted the residual connections and normalization layers. In addition, the decoders incorporated a length regulator to align the gloss and text sequences. [53] implemented a transformer encoder-decoder with bidirectional attention using mBART initialization. [16] employed a decoder-only transformer to translate from gloss to text as part of a S2G2T end-to-end framework. In this work, a direct comparison with an encoder-decoder transformer was not conducted. All these works tested their architectures on PHEONIX-14T dataset. Hence, when compared, we find that [53] outperformed the other two architectures, indicating that the encoder-decoder transformer architecture is the most precise. Other works applied transformer encoder-decoder in its original architecture in T2G [19], G2S [19], [134], and S2T translations [16], [18], [52], [53], [119], [126], [130], [131], [133]. [53] reported that their S2T approach significantly improved compared to S2G2T translation due to the visual pre-training using mBART. However, [16] found that S2T encountered challenges with long-term dependencies. Similarly, [18] and [52] found that despite encountering a bottleneck in the form of gloss representation, the S2G2T approach outperformed S2T for the transformer architecture. [53] outperformed all architectures that tested the algorithm on PHOENIX-2014T dataset [16], [18], [52], [53], [119], [126], [130], [133]. We believe that this is due to the architecture's parameters. Regardless of the different types of transformers, the architecture's core component is the self-attention mechanism [150], which focuses on different parts of original sentences while generating the translated sentences.

Initially, the encoder-decoder transformer architecture was proposed by [150] in 2017 for MT. The process of translation using this algorithm starts with tokenization to convert the original sentence into words or sub-words and build the model's vocabulary. This is followed by transforming these sentences into vectors through an embedding process. Then, these embeddings are encoded through positional encoding to determine the position of each word or the distance between different words in the sequence, as shown in Equations (26) and (27). The output of the positional encoding is then passed through the encoder, along with the embeddings, to produce a sequence of encoded representations that capture the relevant information in the sentence. The goal of adding positional encoding values to the embeddings is to provide meaningful distances between the embedding vectors once they are fed into the encoder. The encoder consists of multi-head self-attention, a feedforward network, and normalization and residual connections. The multi-head self-attention allows the model to weigh the importance of different parts of the input and capture relationships between words by computing weights for multiple sets of Query/Key/Value (Q/K/V) vectors, as presented in Equations (28) and (29). These vectors are obtained by multiplying the input embeddings by different weight matrices. On the other hand, the feedforward network adds non-linearities, helping to capture more complex patterns. Moreover, the normalization and residual connections stabilize and speed up the training of deep networks. The decoder of the transformer generates a sequential output using an autoregressive approach through the encoded representations and the previously generated words as input. These sequences are fed into a linear layer and then a SoftMax layer to output a set of probabilities for the generated word.

$$PE(pos, 2i) = sin(\frac{pos}{1000^{\frac{2i}{d_{model}}}}) \quad (26)$$

$$PE(pos, 2i+1) = cos(\frac{pos}{1000^{\frac{2i}{d_{model}}}}) \quad (27)$$

$$Attention = Sofmax(\frac{QK}{\sqrt{d}})V \quad (28)$$

$$MultiHead = Concat(Attention_1 \ldots Attention_i) * w^o \quad (29)$$

where $pos$ is the current position, $i$ is the dimension index. $d$ is the dimension. $w^o$ is a learned weight matrix that adds the parameters in a single-head self-attention. In summary, Transformers [150] altered the MT task, due to their ability to capture dependencies across sequences and solve the issues faced by previous models. However, this algorithm generates each token sequentially, as it is conditioned on the previously generated token sequence. This process is not parallelizable and is slow during inference. Nevertheless, different types of Transformers exist in SLMT systems, including encoder-only transformers, decoder-only transformers (like GPT [42]), and encoder-decoder transformers [150]. Many works applied the transformer and resulted in significant translation precision. Overall, results showed that the encoder-decoder transformer architecture [53] outperformed all other works.

- *Non-Autoregressive Network:* In 2022, [19] implemented a non-autoregressive network to generate sign videos based on glosses, which contains a length regulator and a spatial-temporal graph convolutional generator to produce the sign videos. This model avoids the sequential translation issue faced by transformers by generating all output tokens in parallel, resulting in significantly lower latency during inference [151]. Although non-autoregressive networks can provide faster results, they may be less accurate when handling complex sentence structures or long-distance dependencies. The trade-off between speed and accuracy is critical when choosing the appropriate model for a given application.

In summary, the applied translation algorithms show that neural machine translation is the most common in the SLMT systems, with the dominance of the transformer architecture. This proves its effectiveness in SLMT due to its ability to capture contextual information. In addition, various performance metrics are implemented to evaluate the models. BLEU-1 and BLEU-4 are the most used, emphasizing that n-gram precision and phrase-level correctness are critical concerns in SLMT. It is worth noting that the bigger the n-gram, the more challenging it is to achieve precision. We also found that PHOENIX-2014 dataset is the most common, as shown in Fig. 10. Therefore, in Fig. 11, we compare the works that tested their models on this dataset. In the T2G translation, the GRU applied in [118] slightly outperformed the encoder-decoder transformer fusion applied in [19]. On the other hand, in G2T translation, the encoder-decoder transformer applied in [53] outperformed the decoder-only transformer applied in [16], the encoder-decoder transformer fusion applied in [19], and the GRU applied in [17]. However, [19] reported that the transformer fusion outperformed the original transformer when tested on the same dataset. Moreover, the encoder-decoder transformer in [119] outperforms the encoder-only transformer in [16] during S2G translation. Similarly, in G2S, the encoder-decoder transformer in [134] outperformed the non-autoregressive network in [19]. The high word error rate here reflects the difficulty in recognizing context. When comparing the architecture applied in S2G and G2S, we notice that the one in [134] is the most effective. We believe that this is due to the architecture parameters. On the other hand, in the S2G2T and S2T translations, the encoder-decoder transformers in [52] and [53] outperform all other architectures and algorithms, respectively. This analysis shows that the encoder-decoder transformer is the most efficient.

However, although the transformer in [53] surpasses all other architectures globally when tested on PHOENIX-2014T dataset, it is not as efficient when tested on CSL-Daily dataset. This is due to the difference in the dataset size, as PHOENIX-2014T dataset [17] is 3.3 times larger than CSL-Daily dataset [52]. Lastly, most works on S2G2T translation [18], [52], [53], [119], [120], [122] and S2T translation [18], [52], [53], [113], [125], [126], [130], [131], [132], [133] consider various sign languages, addressing the need to create multilingual SLMT models.

**Table 4:** Comparison between Sign Language Machine Translation Algorithms.

| Translation | Work | Sign Language | Dataset | Translation Model: Algorithm | Translation Model: Feature Extraction | Performance Metrics: BLEU-1 | BLEU-2 | BLEU-3 | BLEU-4 | ROUGE | Word Error Rate | Accuracy | Execution Time |
|---|---|---|---|---|---|---|---|---|---|---|---|---|---|
| Rule-Based Machine Translation (RBMT) | | | | | | | | | | | | | |
| Text-to-Gloss | [54] | Greek | 1015 sentences | Rule-based | POS tagging and chunk partial parser | 90 | NR | NR | 84 | NR | NR | NR | NR |
| Text-to-Gloss | [109] | Arabic | 600 sentences | Rule-based | Morphological and syntactic analysis | NR | NR | NR | 35 | NR | 0.6 | NR | NR |
| Sign-to-Text | [110] | Brazilian | 69 sentences | Rule-based | POS tagging and text parsing | 64.3 | 42.1 | 29.4 | 21.1 | NR | 58.3 | NR | NR |
| Example-Based Machine Translation (EBMT) | | | | | | | | | | | | | |
| Text-t-Gloss | [111] | Dutch | 561 sentences | EBMT | POS tagging | NR | NR | NR | NR | NR | NR | NR | NR |
| Statistical Machine Translation (SMT) | | | | | | | | | | | | | |
| Sign-to-Text | [112] | Arabic | 30 words | Euclidean Distance | Dynamic skin detector, skin-blob tracking, and cascade boosting algorithm | NR | NR | NR | NR | NR | NR | 97 | NR |
| Sign-to-Text | [113] | Thai and American | 10 words for TSL and 31 words for ASL from BOSTON-50* | Fuzzy KNN | Not reported | NR | NR | NR | NR | NR | NR | 91.4 | NR |
| Text-to-Gloss | [114] | American | 300 sentences | Elastic Net Regularization | n-spectrum weighted word kernel | 28 | NR | NR | NR | NR | NR | NR | 5.9 sec/word |
| Text-to-Gloss | [115] | Marathi | 1200 sentences | Phrase-based | Tokenization, POS tagging, and phrase identification | NR | NR | NR | NR | NR | NR | NR | NR |

| Translation | Work | Sign Language | Dataset | Translation Model: Algorithm | Translation Model: Feature Extraction | Performance Metrics: BLEU-1 | BLEU-2 | BLEU-3 | BLEU-4 | ROUGE | Word Error Rate | Accuracy | Execution Time |
|---|---|---|---|---|---|---|---|---|---|---|---|---|---|
| Neural Machine Translation (NMT) | | | | | | | | | | | | | |
| Sign-to-Gloss | [16] | German | PHOENIX-2014T | Transformer | 3D ResNets | NR | NR | NR | NR | NR | 24.6 | NR | NR |
| | [119] | | | | Self-Mutual Knowledge Distillation model | NR | NR | NR | NR | NR | 19.2 | NR | NR |
| Gloss-to-Text | [16] | | | Transformer | | 48.9 | 36.9 | 29.5 | 24.5 | NR | NR | NR | NR |
| | [17] | | | GRU with attention layer† and LSTM with attention layer | | 44.1 | 31.5 | 23.9 | 19.3 | 45.5 | NR | NR | NR |
| | [19] | | | Transformer | Fast Fourier Transform | 33.7 | 24.8 | 19.2 | 15.8 | 36.5 | NR | NR | NR |
| | [53] | | | | mBART’s SentencePiece | 52.7 | 40 | 32.1 | 26.7 | 52.5 | NR | NR | NR |
| Text-to-Gloss | [19] | German | PHOENIX-2014T | Transformer | Fast Fourier Transform | 47.9 | NR | NR | 14.8 | 51.2 | NR | NR | NR |
| | [116] | Portuguese | 150k sentences | LSTM†, and BiLSTM | Morphological Analysis | 95.6 | 93.4 | 91.5 | 85.1 | NR | NR | NR | NR |
| | [117] | Pakistani | 50k sentences | GRU†, BiGRU, BiLSTM, and LSTM with attention layer | NR | 83.3 | NR | NR | 51.4 | NR | 0.2 | NR | NR |
| | [118] | | | GRU | NR | 50.7 | 32.3 | 21.5 | 15.3 | 48.1 | 4.5 | NR | NR |
| Gloss-to-Sign | [19] | German | PHOENIXNR2014T | Non-autoregressive network | NR | 15.6 | NR | NR | 7.2 | 19.1 | 83.3 | NR | NR |
| | [118] | | | GAN | NR | NR | NR | NR | NR | NR | NR | NR | NR |
| | [134] | | | Transformer | | 23.1 | 14.2 | 10.5 | 8 | 23.5 | 81.8 | NR | NR |
| | [135] | German, Indian and American | PHOENIX-2014T, ISL-CSLTR, and UCF101 | GAN | NR | NR | NR | NR | NR | NR | NR | NR | NR |

| Translation | Work | Sign Language | Dataset | Translation Model: Algorithm | Translation Model: Feature Extraction | Performance Metrics: BLEU-1 | BLEU-2 | BLEU-3 | BLEU-4 | ROUGE | Word Error Rate | Accuracy | Execution Time |
|---|---|---|---|---|---|---|---|---|---|---|---|---|---|
| Sign-to-Gloss-to-Text | [16] | German | PHOENIX-2014T | Transformer | | 48.5 | 35.4 | 27.6 | 22.5 | NR | NR | NR | NR |
| | [17] | | | GRU with attention layer | NR | 43.3 | 30.4 | 22.8 | 18.1 | 43.8 | NR | NR | NR |
| | [18] | Chinese and German | SLR-100, PHOENIX-2014T, and CSL daily* | Transformer | Multi-View Spatial-Temporal Embedding Network | 45.4 | 32.5 | 25.1 | 20.4 | NR | NR | NR | NR |
| | [52] | Chinese and German | CSL-Daily & PHOENIX-2014T* | | NR | 48.6 | 36.1 | 28.5 | 23.5 | 49.4 | NR | NR | NR |
| | [53] | American, German, and Chinese | Kinetics-400 and WLASL for pretraining. PHOENIX-2014T and CSL-Daily* for experiments | | mBART's SentencePiece | 50.3 | 37.4 | 28.1 | 21.5 | 51.4 | NR | NR | NR |
| | [119] | Chinese and German | CSL-Daily* & PHOENIX-2014T | Transformer | SMKD model | 54.4 | 40.3 | 30.5 | 23.8 | 53.1 | NR | NR | NR |
| | [120] | | CSL-Daily, and PHOENIX-2014T* | Hierarchical GNN | Hierarchical spatiotemporal graphs | 45.2 | 34.7 | 27.1 | 22.3 | NR | 19.8 | NR | NR |
| | [121] | German | PHOENIX-2014T | Transformer | CNN | 48.2 | 35.6 | 28 | 23.1 | 49.2 | NR | NR | NR |
| | [122] | Chinese and German | CSL, PHOENIX-2014, & PHOENIX-2014T* | BiLSTM | SMC and TM modules | 47 | 36.1 | 28.7 | 23.7 | 46.7 | NR | NR | NR |

| Translation Model | | | | | | Performance Metrics | | | | | | | |
|---|---|---|---|---|---|---|---|---|---|---|---|---|---|
| Translation | Work | Sign Language | Dataset | Algorithm | Feature Extraction | BLEU-1 | BLEU-2 | BLEU-3 | BLEU-4 | ROUGE | Word Error Rate | Accuracy | Execution Time |
| Sign-to-Text | [16] | German | PHOENIX-2014T | Transformer | 46.6 | 33.7 | 26.2 | 21.3 | NR | 26.2 | NR | NR | |
| | [18] | German and Chinese | SLR-100, PHOENIX-2014T*, and CSL-Daily | Transformer | Multi-View Spatial-Temporal Embedding Network, 3D CNN | 49.6 | 36.5 | 29.1 | 22.5 | NR | 23.2 | NR | NR |
| | [49] | American | 69 words used in 41 sentences | BiLSTM | NR | NR | NR | NR | NR | NR | NR | 95 | 1.4 sec/ sign |
| | [52] | German and Chinese | CSL-Daily & PHOENIX-2014T* | Transformer | NR | 50.8 | 37.8 | 29.7 | 24.3 | 49.5 | 23.9 | NR | NR |
| | [53] | German, and Chinese | PHOENIX-2014T*, CSL-Daily | Transformer | mBART's SentencePiece | 54 | 41.8 | 33.8 | 28.4 | 52.7 | NR | NR | NR |
| | [85] | Chinese | CSL | hierarchical-LSTM with attention layer | 3D CNN | 50.8 | 33 | 20.7 | NR | 50.3 | 0.6 | NR | NR |
| | [89] | Columbian | CoL-SLTD | LSTM | 3D CNN | 30.1 | 12.9 | 7.1 | 4.7 | 30.6 | 88.9 | NR | NR |
| | [123] | American | 7,306 samples covering 56 words and 100 sentences | Hierarchical BiDNN with LSTM | Leap motion sensors | NR | NR | NR | NR | NR | NR | 94.5 | NR |
| | [124] | German | PHOENIX-2014T | BiGRU | Frame stream density compression and CNN | 32.2 | 19.4 | 13.7 | 10.7 | 32.3 | NR | NR | NR |
| | [125] | Brazilian, Indian, and Korean | MINDS-Libras*, INCLUDE, KSL, and LIBRAS-UFOP | | 3D CNN | NR | NR | NR | NR | NR | NR | 91 | NR |
| | [126] | Chinese and German | CSL Daily and PHOENIX-2014T* | Transformer | CNN | 36.7 | 25.4 | 18.9 | 15.2 | 38.9 | 0.7 | NR | NR |
| | [127] | German | PHOENIX-2014T | GRU | Semantic Focus of Interest Network with Face Highlight Module | NR | NR | NR | 10.9 | 34.9 | NR | NR | NR |
| | [128] | Chinese | CSL-Daily | LSTM with attention layer | 3D CNN | 45 | 23.8 | 12.7 | NR | 44.9 | 67.2 | NR | NR |
| | [129] | Malaysian | 19 signs | CNN with attention layer | CNN | NR | NR | NR | NR | NR | NR | 99.4 | NR |
| | [130] | Chinese and German | CSL-Daily and PHOENIX-2014T* | Transformer | CNN | 47.2 | NR | NR | 19.7 | 46.2 | NR | NR | NR |
| | [131] | Multi-language | SP-10 | Transformer | NR | NR | NR | NR | 4.7 | 32.7 | NR | NR | NR |
| | [132] | German and Korean | PHOENIX-2014T and KETI* | GRU | AlphaPose, Normalization, Stochastic Augmentation, and Skip Sampling | NR | NR | NR | 85.5 | 84.9 | NR | NR | NR |
| | [133] | American and German | ASLing and PHOENIX-2014T* | Transformer | CNN, OpenPose, and Optical Flow embeddings | 39.2 | 24.6 | 16.9 | 12.3 | NR | NR | NR | NR |

*: Best dataset performance. †: Best algorithm performance. NR: Not reported.

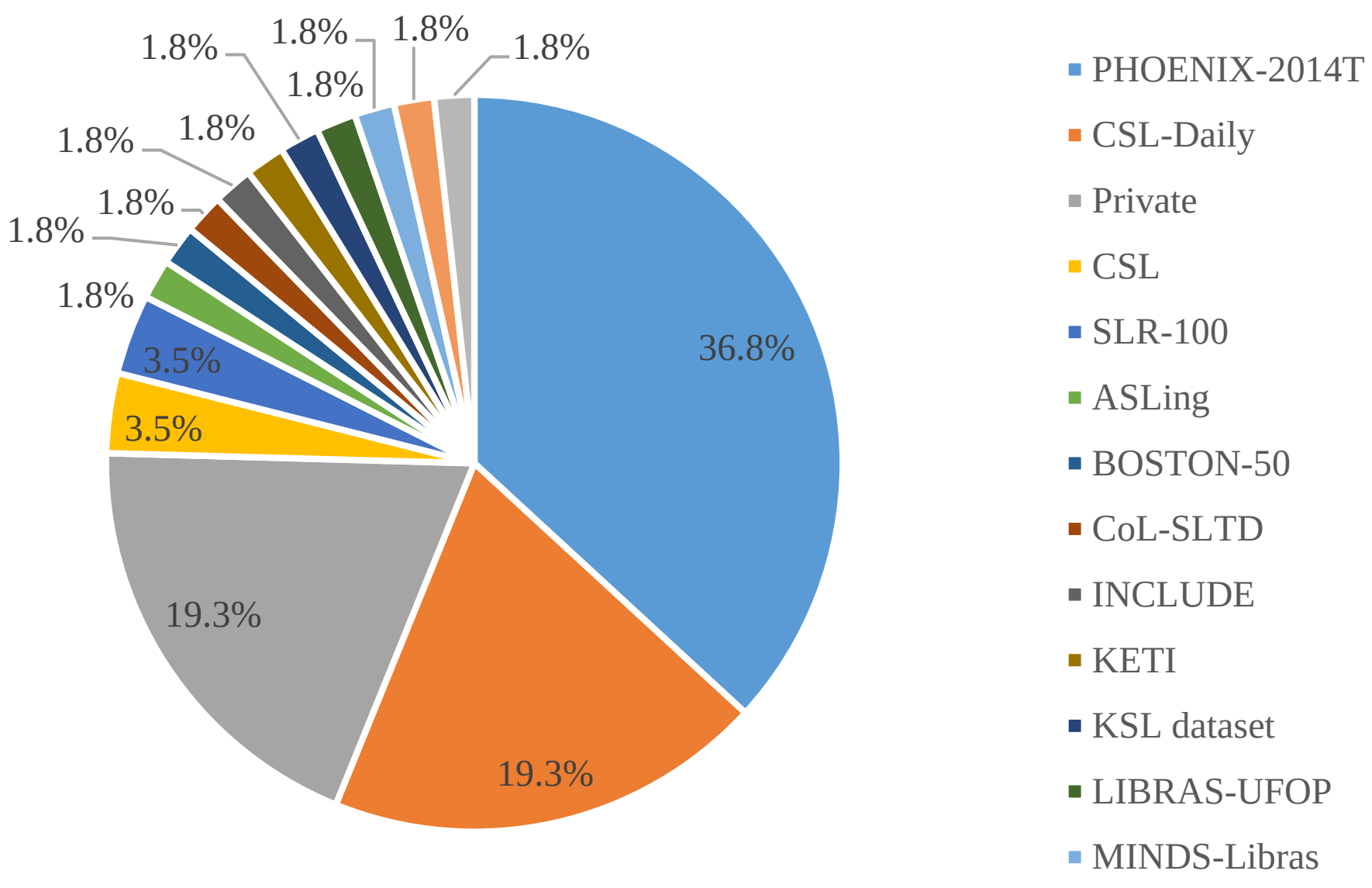

**Fig. 10.** Datasets frequency in literature.

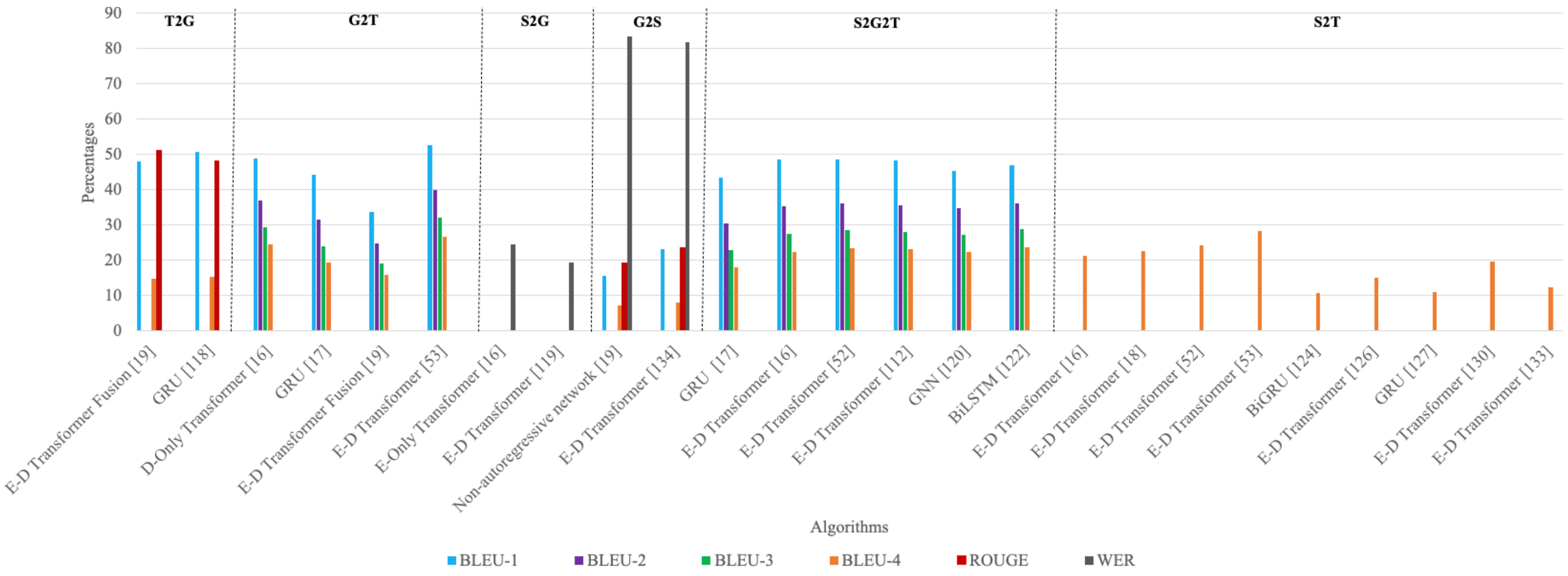

**Fig. 11.** Comparison between Sign Language Machine Translation Algorithms Based on PHEONIX-2014T Dataset. E refers to Encoder, and D refers to Decoder.

# 9 Taxonomy of Transformer Architectures for Sign Language Machine Translation

We present a classification of the component-based Transformer architectures in Fig.12. We classify these architectures into four categories: 1) Encoder-decoder transformer (EDT) (Fig. 12 (a)), 2) Encoder-only transformer (EOT) (Fig. 12(b)), 3) Decoder-only transformer (DOT) (Fig. 12(c)), and 4) Encoder-decoder transformer fusion (EDTF) (Fig. 12(d)). Table 5 presents a comparison between the works that used these transformer architectures, along with their parameters, in the different stages of our proposed framework.

The EDT architecture was proposed by [150] in 2017. The architecture consists of a *Tokenization* block to transform the input sentences into words or sub-words that build the model's vocabulary. In addition, the architecture contains an *Embedding* block to transform the input sentences into vectors, plus a *Positional Encoding* block to provide information about the position of each word in the sentence. The model also consists of *Encoder* and *Decoder* layers. The encoder consists of the following components: *Multi-Head Attention* to weigh the importance of different parts of the input and capture the relationships between the input words. *Feedforward Neural Networks* add non-linearities and help the model capture more complex patterns. *Normalization and Residual Connections Layers* to stabilize and speed up the training of deep networks. The decoder consists of components similar to the encoder and a *Masked Multi-Head Attention* to prevent the model from accessing information about future positions. Following the decoder, the model consists of a *Linear Layer* to transform the decoder output into the desired dimension and a *SoftMax* activation to obtain the probability distribution over the vocabulary for each position in the output. This architecture has been widely used across various translation stages, including S2G [53], [119], G2T [53], G2S [134], S2G2T [18], and S2T [18], [52], [53], [126], [130], [134].

The EOT and DOT can be applied in a model that utilizes other algorithms. For instance, the GPT architecture consists of a Byte Per Encoding and a decoder-only transformer [42]. However, in the SLMT literature, the encoder-only and decoder-only transformers were applied in one work [16]. The researchers here applied the encoder for the S2G translation. The encoder was trained using a CTC loss to predict the gloss sequences. They also applied the decoder for the G2T translation. They trained and tested each part independently and joined them in an end-to-end S2G2T SLMT system.

Furthermore, [19] implemented a variation of the encoder-decoder transformer that we call an "encoder-decoder transformer fusion" (EDTF) for G2T, T2G, and G2S translations. The model includes a Fast Furrier Transform (FFT) block between the embedding layer and the encoder. This is added to extract each word's frequency representation [152] as an additional feature to the embeddings. To enhance the gloss translation, the researchers incorporated a gated bilateral fusion layer within the decoder between the self-attention and the feed-forward network. This mechanism controls the influence of the input and its context while controlling the gradients of the propagation [153]. In addition, they added a CTC joint training layer after the encoder. CTC identifies the repetition of the output and blank outputs and aims to rectify the alignment between the input and output; hence, it allows the model to converge early [154]. Moreover, they excluded the residual connections and normalization layers in their architecture.

In summary, the Transformer is the most common algorithm in SLMT systems. Overall, the parameters of the encoder-decoder transformer applied in [53] are the most effective. However, no works compare all four types of transformers in each translation phase i.e. S2G, G2T, T2G, G2S, S2T, and S2G2T translations. This gap should be filled in future works to find the most efficient and accurate SLMT system.

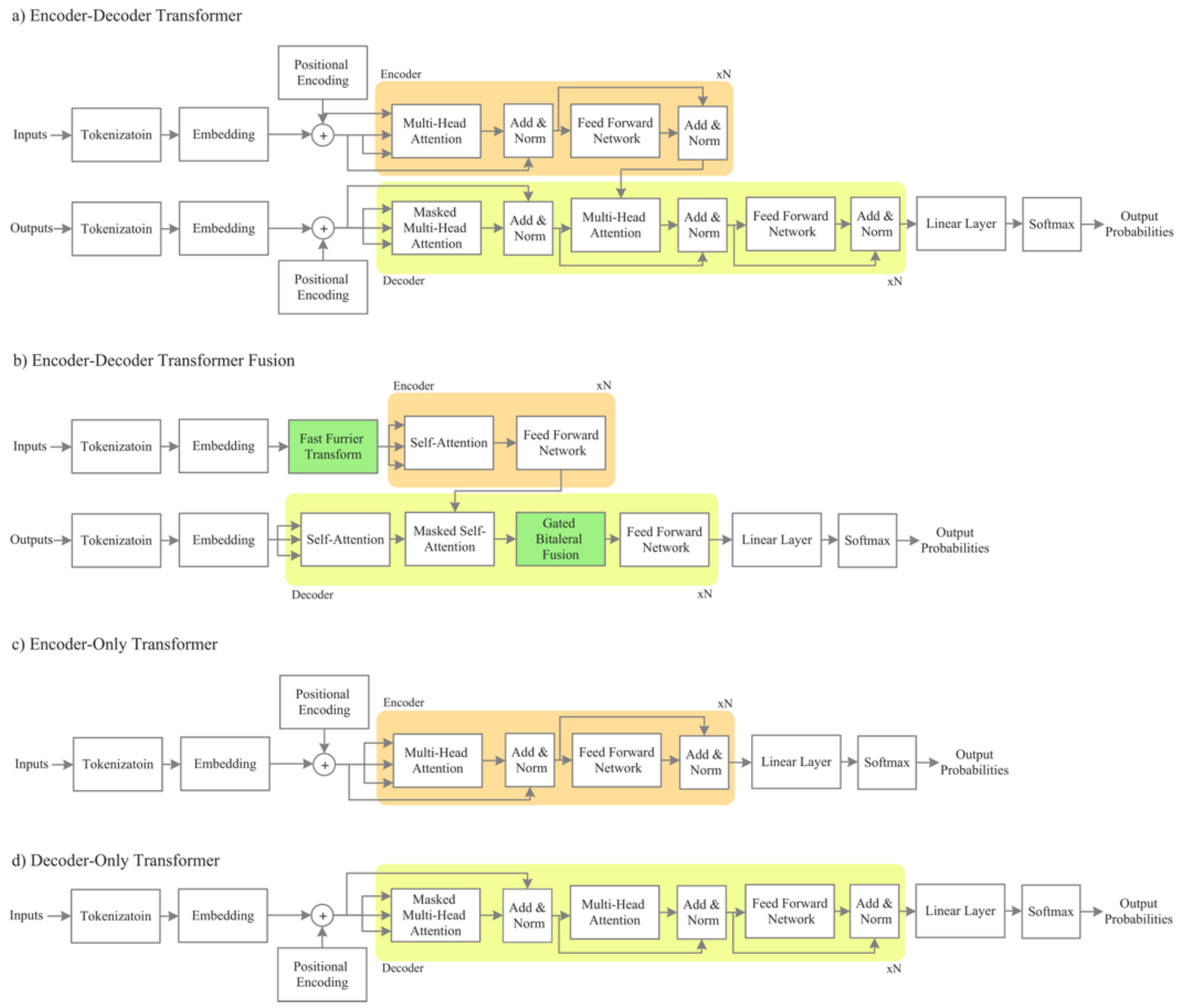

**Fig. 12.** Taxonomy of Transformer architectures.

**Table 5:** Hyperparameters of the Transformer architectures.

| | | SLMT Covered | | | | | | Architecture | | | | | | | | | | | |
|---|---|---|---|---|---|---|---|---|---|---|---|---|---|---|---|---|---|---|---|
| Work | Algorithm | Sign-to-Gloss | Gloss-to-Text | Text-to-Gloss | Gloss-to-Sign | Sign-to-Gloss-to-Text | Sign-to-Text | Number of encoders | Number of decoders | Hidden units | Number of heads | Number of attention blocks | Optimizer | Dropout | Learning rate | Epochs | Number of Iterations | Batch Size | Weight decay |
| [19] | EDTF | ✗ | ✓ | ✓ | ✓ | ✗ | ✗ | NR | NR | 256 | 8 | 2 | Adam | NR | 0.001 | NR | NR | NR | NR |
| [53] | EDT | ✓ | ✓ | ✗ | ✗ | ✗ | ✓ | 12 | 12 | 1024 | 16 | NR | NR | NR | NR | NR | NR | NR | NR |
| [16] | EOT and DOT | ✓ | ✓ | ✗ | ✗ | ✓ | ✗ | NR | NR | 512 | 8 | NR | Adam | 0.1 | $10^{-6}$ | NR | 100 | 32 | $10^{-3}$ |
| [119] | EDT | ✓ | ✗ | ✗ | ✗ | ✗ | ✗ | 2 | 2 | 512 | 8 | NR | Adam | 0.4 | NR | NR | NR | NR | NR |
| [18] | EDT | ✗ | ✗ | ✗ | ✗ | ✓ | ✓ | 3 | 3 | 512 | 8 | NR | Adam | 0.3 and 0.6 | $10^{-7}$ | 100 | NR | 32 | $10^{-3}$ |
| [52] | EDT | ✗ | ✗ | ✗ | ✗ | ✗ | ✓ | NR | NR | 512 | 8 | NR | Adam | 0.1 | $5 \times 10^{-5}$ | 30 | NR | 32 | $10^{-6}$ |
| [126] | EDT | ✗ | ✗ | ✗ | ✗ | ✗ | ✓ | 3 | 6 | NR | NR | NR | Adam | NR | $3 \times 10^{-5}$ | 300 | NR | 100 | NR |
| [130] | EDT | ✗ | ✗ | ✗ | ✗ | ✗ | ✓ | 1 | 1 | 512 | 8 | 1 | Adam | NR | $5 \times 10^{-4}$ | NR | NR | 4 | NR |
| [133] | EDT | ✗ | ✗ | ✗ | ✗ | ✗ | ✓ | 3 | 3 | 512 | 8 | NR | Adam | 0.001 | NR | 70-150 | NR | NR | 0.001 |
| [134] | EDT | ✗ | ✗ | ✗ | ✓ | ✗ | ✗ | 2 | 2 | 512 | 4 | NR | Adam | NR | $10^{-3}$ | NR | NR | NR | NR |

EDT: Encoder-Decoder Transformer, EOT: Encoder-Only Transformer, DOT: Decoder-Only Transformer, EDTF: Encoder-Decoder Transformer Fusion, NR: Not Reported.

# 10 Performance Evaluation of Transformers for Gloss-to-Text Translation: A Case Study

To understand the implications of deploying transformers-based sign language interpretation in real scenarios, we conduct empirical evaluations of the four transformer architectures which are underpinned by our taxonomy, in a unified environmental setup. It is necessary to deploy an efficient transformer in real-world scenarios. In this case study, we consider sign language gloss to spoken language text (G2T) machine learning translation.

## 10.1 Datasets

We use the largest publicly available sign language PHOENIX-2014T dataset [17]. It consists of 8,257 weather-related sentences in German sign language. To assess the impact of a small dataset on the comparative performance of the different transformer architectures, we employ random 500 unique sentences of PHOENIX-2014T dataset. To evaluate the performance on a different type of dataset and language, we collect our own private ASL dataset, which we call "*MedASL*". It consists of 500 medical-related sentences generated by ChatGPT [155] to reflect scenarios between patients and doctors, nurses, technicians, and registration desk staff in a medical center. An ASL expert then translated these sentences to gloss.

## 10.2 Experimental Setup

To achieve the best performance possible in a sign language interpretation scenario, we perform hyperparameter tuning which determines the optimal values for transformer models' parameters. The parameters we study for each transformer model are presented in Table 6. These parameters are selected as the best parameters reported in the state of the art as described in Table 5. Consequently, for each dataset, we perform 16 experiments, corresponding to each combination of the four architectures and their respective hyperparameters configurations.

**Table 6:** Hyperparameters in the literature for the Transformer architectures under study.

| Algorithm | Number of encoders | Number of decoders | Hidden units | Number of heads | Number of attention blocks | Optimizer | Dropout | Learning rate | Batch size | Weight decay | Hyperparameter tuning | Justification for hyperparameter selection | Assigned name for hyperparameters combination |
|---|---|---|---|---|---|---|---|---|---|---|---|---|---|
| EDT [19] | NR, 1* | NR, 1* | 256 | 8 | 2 | Adam | NR, 0* | 0.001 | 32 | NR, 0* | ✗ | NR | H1 |
| EDTF [53] | 12 | 12 | 1024 | 16 | NR, 1* | NR, Adam* | NR, 0* | 0.001 | NR, 32† | NR, 0* | ✗ | NR | H2 |
| | | | | | | | | $10^{-6}$ | | | | | H3 |
| DOT [16] | NR, 1* | NR, 1* | 512 | 8 | NR, 1* | Adam | 0.1 | $10^{-6}$ | 32 | $10^{-3}$ | ✗ | NR | H4 |

EDT: Encoder-Decoder Transformer, EDTF: Encoder-Decoder Transformer Fusion, EOT: Encoder-Only Transformer, DOT: Decoder-Only Transformer, H1: Hyperparameters configuration 1, H2: Hyperparameters configuration 2, H3: Hyperparameters configuration 3, H4: Hyperparameters configuration 4, NR: Not Reported, *: Default value as it was not reported in literature; † : The same as the optimal value in other architectures

## 10.3 Experimental Results Analysis

Fig. 13 shows our results for the four transformer architectures with the best hyperparameters configuration. EDT architecture has the best performance across all datasets. This reveals that EDT maintains performance consistency regardless of the dataset size. This is because the encoder captures the gloss context then the decoder produces the translated text more efficiently. Such architecture is crucial for sequence-to-sequence tasks like SLMT. In contrast, EDTF has the worst performance. This is due to omitting the positional encoding and normalization layers, which leads to translation inefficiency. This underperformance indicates fundamental limitations in the integration of the encoder and decoder networks or incompatibility of the fusion approach with G2T translation. DOT performed better than EOT across all datasets. This is due to the capability of the decoder to generate sequences token by token based on previous tokens. However, EOT is not suitable for translating sequences and focuses on input understanding and representation.

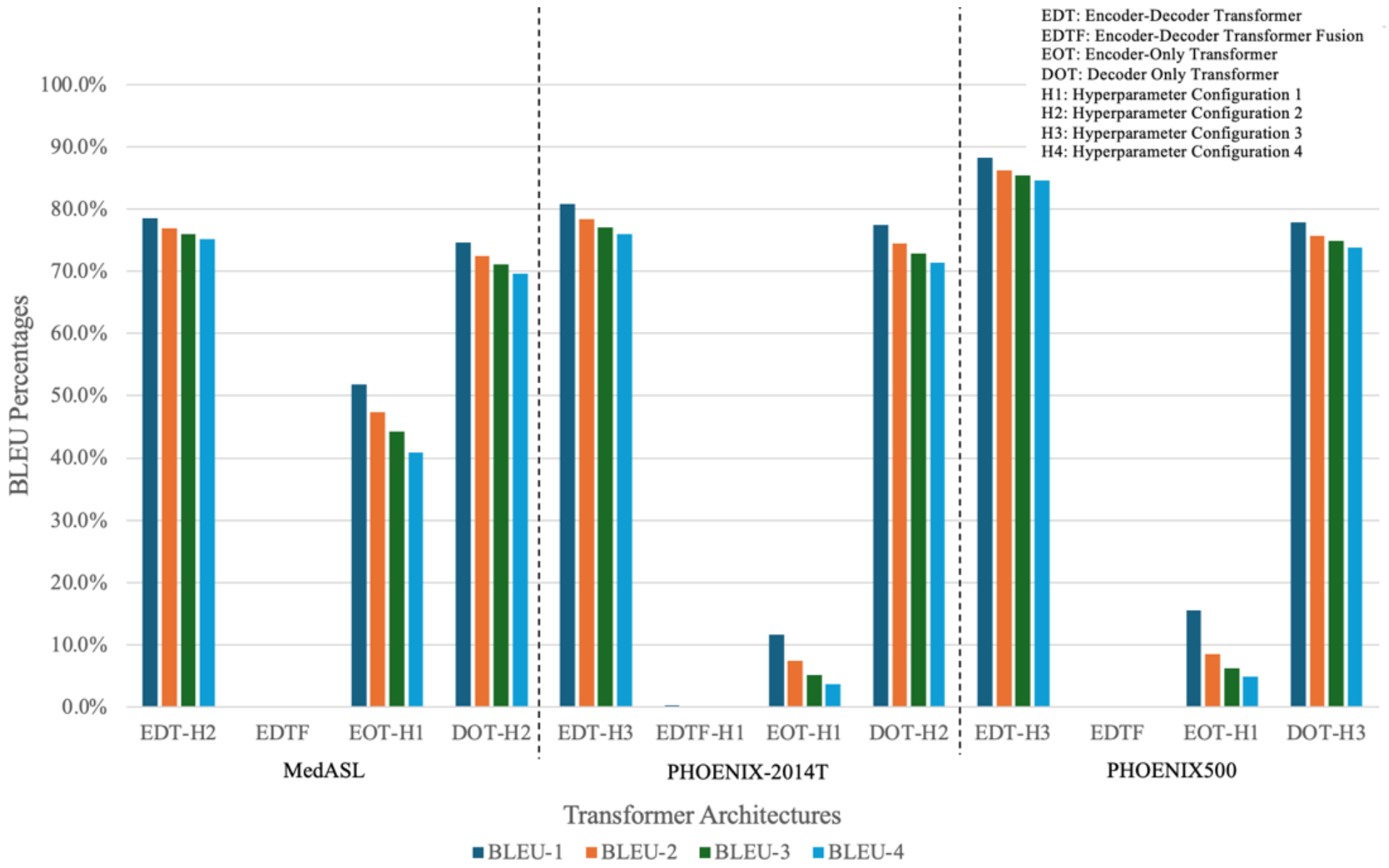

**Fig. 13.** Comparison between the four transformer architectures for gloss-to-text translation with the best hyperparameter configuration on MedASL, PHOENIX-2014T, and PHOENIX500 datasets.

In summary, for SLMT G2T translation, EDT outperforms all other architectures due to its capability to combine context understanding of gloss and sequence-to-sequence translation. Regarding hyperparameter tuning, our results reveal that EDT and DOT architectures' performance increased with the increasing number of heads and hidden units.

## 11 Challenges and Proposed Solutions

A real-time efficient SLMT system comprises a user-centric architecture that allows DHH and hearing users to interact with each other through a simultaneous translation from sign to spoken language and vice versa. The challenges of such a system include:

- **Availability of large and multilingual datasets:** There are more than 300 sign languages globally [3], many of which are low in resources, such as the ArSL [156]. Consequently, the number of available datasets is scarce. Therefore, researchers need a large and accurate dataset for faster data generation and precision [51]. In addition, sign language datasets often include the translation between one spoken language and one sign language (e.g., Chinese and Chinese Sign Language) [21]. Consequently, limitations of current technologies include the training of monolingual AI models, in which models are usually trained on one pair of languages at a time. A solution for this challenge is the creation of crowdsourced platforms [157] for collecting, annotating, and translating sign language data. This approach overcomes the scarcity of resources in a particular sign language. It leads to the creation of a large multilingual dataset, which reduces the need for extensive datasets for each language. Another solution is to apply transfer learning techniques by pre-training a model on rich sign language datasets and then fine-tuning them on low-resource sign language [53]. This strategy is particularly relevant for including regional dialects and idiomatic expressions, which are critical for the accurate representation of sign languages in AI models.
- **High deployment cost:** The deployment of SLMT systems involves substantial financial investment, due to the advanced technology required for accurate sign language recognition and translation and the need for extensive infrastructure, including servers for processing and storing large datasets. The economic barrier extends to the cost of developing, testing, and continually updating these systems to accommodate new sign languages and dialects, making it a significant obstacle to widespread adoption. A potential solution to mitigate these costs is utilizing cloud and edge computing to reduce the need for expensive, dedicated infrastructure [158], and by distributing the computational load more efficiently and cost-effectively across edge and cloud servers [159].
- **High retraining cost:** Languages are constantly evolving [160]. Therefore, the current datasets will soon need to be updated, and current AI models must be retrained. This leads to high electricity and maintenance costs. [53] applied transfer learning techniques to grasp the knowledge from rich sign languages and apply it to low-resource languages. Through this approach, the model could learn universal sign language features, including the translation of idiomatic expressions and regional dialects. On the other hand, domain adaptation strategies allow the AI model to adjust to a specific in-domain low-resource sign language and out-of-domain translations for the same sign language with high translation accuracy [161].
- **High energy consumption:** SLMT systems involve Internet-of-Things (IoT) devices and edge and cloud servers to train the AI models and perform real-time translation. These components consume a high amount of energy. By 2025, global data centers will consume around 800 terawatt-hours of electrical energy, triple the consumption in 2021 [162]. A solution for such a challenge is the introduction of energy-aware strategies for resource distribution and allocation [163].

- **QoS and latency management:** An ultra-low latency is essential yet challenging for a real-time simultaneous translation, especially in critical situations. In the case of SLMT, sentences in sign languages have different structures from spoken languages [4], making it challenging to start the translation before a sentence ends. [164] proposed a solution to apply reinforcement learning for sequence-level optimization in addition to multi-task learning, which allows a model to be trained on multiple tasks simultaneously, such as POS tagging and machine translation. This solution facilitates domain adaptation and transfer learning, leading to faster training convergence, improved translation quality, and reduced latency. Another solution is to train the dataset on the Cloud while processing the lightweight translation on Edge servers near the translation device [56]. However, to our knowledge, no work provides a comparative analysis between a human interpreter and SLMT system in terms of execution time, leaving a gap in determining the efficiency of SLMT systems in real-world scenarios. Conducting such a study is necessary to provide a clearer picture of current SLMT system capabilities and areas where they still lag behind human performance.
- **Data privacy and security:** Security mechanisms should be in place for SLMT systems as the translation exposes personal information and surroundings during video transmission [165]. Therefore, intelligent approaches are required to automate security responses in the IoT network. Security threats in such scenarios include API attacks, eavesdropping, and DDoS attacks can be mitigated through intrusion detection and digital forensics techniques [57]. In gathering sign language data, collecting consent forms from system users is crucial for DHH individual privacy. In a peer-to-peer environment where the systems are used by multiple stakeholders, such as healthcare professionals, in communication with the DHH individual, it is very important to ensure the integrity of the consent forms. A promising approach is the employment of blockchain technology that enables a privacy-preserving and secure environment among peers without intermediate third-party using the consensus protocol, and the reinforcement of consent rules via the execution of blockchain smart contracts [166]. This method not only provides a secured record of consent but also empowers users with the ability to grant, modify, or transparently withdraw their consent, thus offering an additional layer of security and trust in the consent process.
- **User acceptance:** Achieving widespread user acceptance SLMT systems presents a significant challenge. A survey by [167] showed that demographics, such as educational background, and technology experiences significantly influence acceptance of automated SLMT systems. Furthermore, it revealed frequent sign language users tend to be more critical of the automated systems due to higher QoS expectations and authenticity of sign language representation. The general skepticism towards automated SLMT systems as replacements for live interpreters or in telephone relay services further complicates the acceptance, as participants are concerned about the system's ability to convey complex sign language nuances accurately and effectively [168]. To overcome this challenge, we propose adopting a participatory design approach that involves DHH users in the development process. This is to ensure that the SLMT systems are tailored to meet the diverse needs and expectations of the DHH community.

## 12 Future Research Directions

In this section, we identify several promising directions for future research in this field. A critical area of focus is the development of more robust and scalable end-to-end SLMT systems that can handle a wide range of sign languages efficiently in terms of precision and execution time. While Transformer is the most used approach for sign language translation, as revealed by our survey, Reinforcement learning would aid in leveraging Transformer translation [169] for sign language translation. In addition, integrating sign language features, such as facial expressions and body language, could further enhance the accuracy and naturalness of SLMT systems.

Transformer models have shown exceptional ability in understanding and generating complex spoken language patterns [150], which could be extended to the nuances of sign languages, with the potential of increasing the translation precision of an LSMT system. However, training the Transformer large datasets suffers from performance efficiency in terms of execution time, which hinders its deployment in real-time SLMT systems. Quantum computing is a promising technology to boost the performance of computing power-hungry applications [170]. It could reduce the computational time required for training deep learning models on large datasets due to their processing power. Therefore, the development of end-to-end SLMT systems enabled by quantum computing would revolutionize the domain of sign language translation, toward the deployment of more sophisticated SLMT systems while increasing their precision and execution time. This would encourage the adoption of these systems by DHH and hearing populations, enabling real-time conversation.

However, deploying a real-time conversation end-to-end SLMT system faces privacy and security challenges of conversational data and videos collected for the DHH individuals. Blockchain has shown its potential in privacy-preserving health records management systems [166]. by creating decentralized ledgers to manage and securely store DHH individual data and consent logs, blockchain technology can provide a robust framework for protecting sensitive information while maintaining DHH individual trust in the SLMT system [171].

Furthermore, to provide the users of SLMT systems with immersive, interactive, and engaging experiences for their users, the integration of scalable and real-time Metaverse [56], along with Virtual Reality (VR) [172], Augmented Reality (AR) [173], wearable devices [174], and educational tools [175] holds promise for revolutionizing how the DHH community interacts with the world.

## 13 Conclusion

The increasing number of Deaf and Hard of Hearing (DHH) worldwide with limited certified sign language interpreters has led to a need for an efficient, signs-driven, integrated end-to-end automated Sign Language Machine Translation (SLMT) system. Many works on the topic gained attention in recent years. Most of the works in the literature proposed systems based on Neural Machine Translation (NMT) algorithms to achieve accurate translation. Our study identifies different Transformer architectures as the most used and effective algorithms in this domain, with PHEONIX-2014T dataset being the most widely applied.

Through this study, we provide an in-depth analysis of the evolution and current state of Machine Translation (MT) systems. We also reveal key insights, challenges, and future research directions, addressing gaps left by existing surveys that have typically focused on specific aspects of sign language translations, such as signs to glosses, glosses to spoken language text, text to glosses, or glosses to signs. In addition, we propose an end-to-end SLMT framework covering all translation stages, i.e. from sign to gloss to text and backward. This framework allows SLMT researchers and developers to systematically evaluate translation methodologies and ensures a comprehensive understanding of the influence of each stage on the overall effectiveness of the translation system. To our knowledge, this is the first work that offers a comprehensive retrospective analysis of the evolution of SLMT algorithms and introduces a taxonomy of Transformer architectures tailored for SLMT. In addition, it presents a Transformer-based gloss-to-text translation case study, comparing the performance of different architectures through empirical evaluations.

One of the key lessons learned is that developing accurate and efficient SLMT systems requires a deep understanding of the unique linguistic features of sign languages. Our analysis demonstrates that Transformer-based architectures surpass other MT models, underscoring the necessity for expansive, annotated datasets and diverse evaluation metrics for thorough SLMT assessment. Moreover, developing accurate and efficient SLMT systems requires a deep understanding of these linguistic features, and the application of large-scale annotated datasets and multiple evaluation metrics, such as BLEU and ROUGE, is critical for comprehensive system assessment.

By addressing these gaps and highlighting the temporal progression of algorithms and the distinct functionalities of Transformer architectures, our work aims to significantly advance the field of SLMT, providing valuable insights and resources that will aid researchers and developers in building more accurate, efficient, and inclusive end-to-end translation systems for seamless communication between the DHH community and the broader society.

## Declarations

This work was supported by the Emirates Center for Mobility Research of the United Arab Emirates University.